\documentclass[10pt,journal,compsoc]{IEEEtran}
\usepackage{amsmath,amsfonts}
\usepackage{algorithmic}
\usepackage{algorithm}
\usepackage{array}
\usepackage[caption=false,font=normalsize,labelfont=sf,textfont=sf]{subfig}
\usepackage{textcomp}
\usepackage{stfloats}
\usepackage{url}
\usepackage{verbatim}
\usepackage{graphicx}
\usepackage{cite}
\usepackage{booktabs}
\usepackage{hyperref}
\usepackage{xurl}

\begin{document}

\title{A survey and classification of face alignment methods based on face models}

\author{
    \IEEEauthorblockN{Jagmohan Meher\IEEEauthorrefmark{1}, Hector Allende-Cid\IEEEauthorrefmark{2}\IEEEauthorrefmark{3}, Torbj{\"o}rn E. M. Nordling\IEEEauthorrefmark{1}}\\
    \IEEEauthorblockA{\IEEEauthorrefmark{1}National Cheng Kung University, Tainan, Taiwan
    \\\{jagmohan.meher, torbjörn.nordling\}@nordlinglab.org} \\
    \IEEEauthorblockA{\IEEEauthorrefmark{2}Fraunhofer IAIS, Germany} \\
     \IEEEauthorblockA{\IEEEauthorrefmark{3}Pontificia Universidad Católica de Valparaíso, Chile
    \\hector.allende@pucv.cl}
}





\maketitle

\begin{abstract}
A face model is a mathematical representation of the distinct features of a human face. 
Traditionally, face models were built using a set of fiducial points or landmarks, each point ideally located on a facial feature, i.e., corner of the eye, tip of the nose, etc. 
Face alignment is the process of fitting the landmarks in a face model to the respective ground truth positions in an input image containing a face. Despite significant research on face alignment in the past decades, no review analyses various face models used in the literature.  
Catering to three types of readers - beginners, practitioners and researchers in face alignment, we provide a comprehensive analysis of different face models used for face alignment. We include the interpretation and training of the face models along with the examples of fitting the face model to a new face image. 
We found that 3D-based face models are preferred in cases of extreme face pose, whereas deep learning-based methods often use heatmaps. 
Moreover, we discuss the possible future directions of face models in the field of face alignment.
\end{abstract}


\section{Introduction}
\IEEEPARstart For all the computer vision problems related to the face, the first step is to detect a face in the given input image or video.
To perform any task on the detected face, it is important to know the exact location of the facial feature, for example, the center of the eye, the corners of the eyes, the tip of the nose, the corners of the lips, the tips of the eyebrows, etc. 
Traditionally, fiducial points (or landmarks) are used on the input face image to locate a specific facial feature.
A set of $n$ number of distinct landmarks can be used to define the form of the face, which constitutes a face model. 
Face alignment refers to the process of fitting the face model to the detected face in an image, that is, moving the $n$ landmark points to localise each landmark on their respective ground truth positions. 
Over the past decades, multiple methods have been proposed to perform this task as it is vital for all face-based computer vision applications. 

The importance of face alignment lies in its ability to enhance the accuracy and reliability of various computer vision and image processing tasks related to the face. For example, it is crucial in face recognition systems \cite{taigman2014deepface}, face expression analysis \cite{zhao2015joint}, face animation \cite{saragih2011deformable}, and biomedical applications like remote photoplethysmography \cite{wang2016algorithmic}. In all of these applications, accurate alignment of face landmarks greatly improves the performance of the underlying algorithms. 

\begin{figure}[htb]
      \centering
          \includegraphics[width=1\columnwidth]{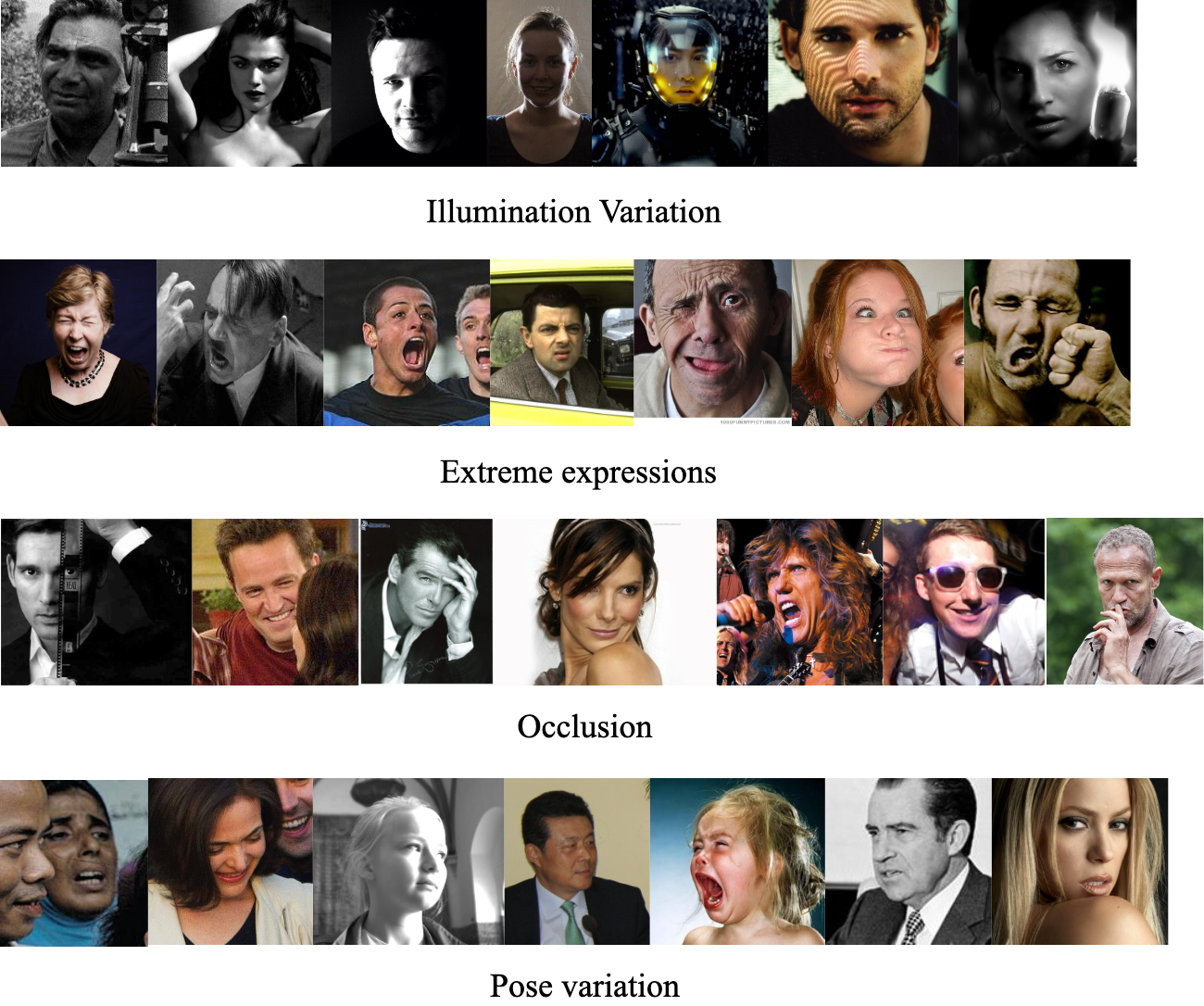}
      \caption{Challenges in face alignment}
      \label{fig:face_alignment_challenges}
\end{figure}

Over the years, significant advancements have been made in the domain of face alignment, with numerous methodologies being introduced, from conventional feature-based techniques to more contemporary deep learning-based approaches. These methods have been evaluated using various benchmark datasets, such as the 300-W dataset, demonstrating varying accuracy. Several factors impact the accuracy of face alignment:
\begin{enumerate}
    \item \textit{Occlusion}: Obstructions in the face region, such as glasses, hair, or masks can hinder the detection of face landmarks. This challenge has been addressed in various studies, for example, the work of \cite{jourabloo2015PIFA}, who proposed a method to handle occlusions using robust features.
    \item \textit{Pose}: Extreme head poses or rotations can make it difficult for face alignment algorithms to accurately identify the face landmarks. In response to this issue, researchers have developed methods like 3D face alignment techniques which align a 3D face mesh model to an input face image. For example, \cite{Zhu2019} proposed a method that addresses pose variations effectively.
    \item \textit{Illumination}: Poor or varying lighting conditions can affect the performance of face alignment algorithms, as it can alter the appearance of face features. \cite{asthana2013robust} proposed a learning-based approach to handle changes in illumination and improve face alignment performance.
    \item \textit{Face expression}: The presence of different face expressions can cause significant variation in the appearance of face landmarks, making it challenging for face alignment methods to accurately detect them. \cite{peng2018jointly} proposed a deep learning-based approach for face landmark detection which is robust even in the cases of extreme expression variations.
\end{enumerate}
Fig. \ref{fig:face_alignment_challenges} shows a few examples of the various challenges in face alignment as mentioned above.
Some of these challenges have been addressed by a couple of the studies. However, there is still room for improvement in the accuracy, robustness, and reliability of face alignment under variable or extreme environmental conditions. 

Several surveys on face alignment and landmark localisation have been published in the last decade \cite{gogic2021regression} \cite{jin2017face} \cite{gao2021survey} \cite{zhang2019advances} \cite{jin2022review}. The survey \cite{jin2017face} provides an overview of various face alignment methods before 2017. \cite{gogic2021regression} focuses on regression-based face alignment methods. \cite{gao2021survey} is a short survey on deep learning-based face alignment methods. \cite{zhang2019advances} talks about video-based face alignment and \cite{jin2022review} about traditional face alignment methods like ASM and AAM. 
All these survey papers briefly talk about the face models used in face alignment. 
However, there is no survey paper yet, that analyses and compares all the face models used for face alignment. 
Our motivation to write a survey paper is to perform a systematic review of the face models used in face alignment. 
We intend to present an in-depth analysis of different perspectives of face alignment that will help a reader of any caliber to understand the concepts and methods used for face alignment. 

We follow the Preferred Reporting Items for Systematic Reviews and Meta-Analyses (PRISMA) to conduct this review \cite{page2021prisma}. 
PRISMA is an evidence-based framework that provides guidelines for conducting and reporting systematic reviews and meta-analyses in scientific research. PRISMA aims to improve the transparency, consistency, and quality of these types of studies. 

In this survey article, we focus on discussing various face models that have been used for effective face alignment.
We highlight the importance of each face model by examining relevant research papers that use the face models along with fitting algorithms for face alignment. 
Additionally, we provide insights from papers that describe these face models and their applications.
To serve as a one-stop resource for face alignment, we also include various open-source implementations of these face models used for face alignment. 
Our goal is to provide a comprehensive overview to a diverse range of readers, including beginners, practitioners, and researchers interested in face alignment. 
For beginners, it serves as an introduction to the topic, offering an accessible starting point to understand the concepts and techniques involved. 
Practitioners can benefit from the practical resources and open-source implementations listed in this review. 
Researchers will find valuable insights from the detailed explanation of training a face model from scratch which can be used along with  fitting algorithms to perform face alignment. 
We will publish our implemented codes in \url{https://github.com/nordlinglab/FaceAlignment-Survey}. 

The major contributions of this paper are:
\begin{enumerate}
    \item  We provide four different perspectives on classification of the face alignment methods. 
    \item  We provide an overview and an in-depth comparison of the face models used in face alignment. 
    \item  We discuss the potential future directions for the usage of face models in face alignment. 
\end{enumerate}

This work is structured as follows. In Section \ref{section:classify}, we present four different perspectives or ways that face alignment methods can be classified. In this paper, we focus on the Output Representation of the face. Thus, in Section \ref{section:face_models}, we elaborate on various mathematical models of a face used in face alignment. It involves understanding the mathematical equations used to define a face. We leave details on the face fitting process in face alignment for the second part of this survey. In section \ref{section:Discussion}, we compare all face models and analyse the pros and cons, and the various applications of each face model. Finally, in Section \ref{section:Conclusion} we conclude the first part of the survey by providing future directions for the use of face models for face alignment.

\section{Different ways to classify face alignment methods} \label{section:classify}
\begin{figure}[htb]
      \centering
          \includegraphics[width=1\columnwidth]{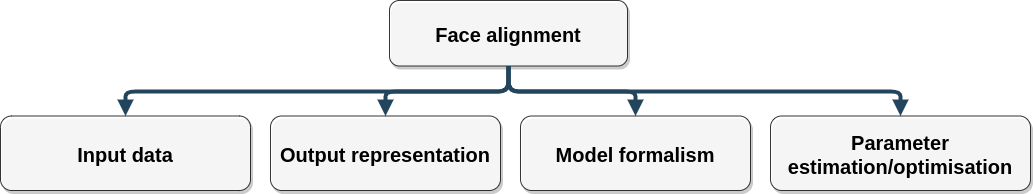}
      \caption{Different ways to classify the face alignment methods}
      \label{fig:face_alignment_categorisations}
\end{figure}

As presented in Fig. \ref{fig:face_alignment_categorisations}, face alignment methods can be classified based on:
\begin{enumerate}
    \item Input data, e.g. use of information only from the current image/frame or also from some past frame in a video
    \item Output, i.e. the representation of the face, such as heatmap or landmarks/key points
    \item Model formalism, e.g. artificial neural network
    \item Parameter estimation/optimisation method, e.g. gradient descent, t is in many cases hard to distinguish between model formalism and parameter optimisation, since one is used intertwined with the other.
\end{enumerate}

\subsection{Based on input data}
Input data is a crucial factor in face alignment methods, as it determines the type and amount of information available for processing. 
The input data can be categorized based on the source of the information, such as a single image, a sequence of images from the same camera, or synchronized images from multiple cameras, and the metadata required such as a bounding box of each person/face.

\textit{Single image:} In this category, face alignment methods work with only one image at a time. The main challenge here is the limited amount of information available, as the methods need to handle variations in pose, expression, lighting, and occlusion with just one sample. 
Active Appearance Models (AAMs) \cite{cootes2001active} is a generative model that learns the appearance variations from a training set and optimises the parameters to fit a single input image.
Constrained Local Models (CLMs) \cite{Cristinacce2006CLM} is a discriminative model that uses local feature detectors learned through regression, iteratively updating shape parameters based on a single image input.
Explicit Shape Regression (ESR) \cite{Cao2012ESR} is a regression-based method that learns a cascade of models to estimate face landmarks' positions from a single image's features.
Deep Alignment Network (DAN) \cite{kowalski2017DAN} is a convolution neural network-based method that predicts face landmarks in a coarse to fine manner using a stacked architecture, refining landmark predictions based on a single input image.

\textit{A sequence of images from the same camera:} In this category, face alignment methods uses a series of images captured by the same camera. The additional temporal information can help improve the alignment, as it provides more context and helps overcome challenges present in single-image methods. 
3D Morphable Models (3DMMs) \cite{blanz19993DMM} are generative models that represent faces as a linear combination of shape and texture basis vectors. Multiple frames help estimate the shape, pose, and texture parameters for better alignment.
Recurrent Neural Networks (RNNs) exploit temporal information in image sequences, modeling temporal dependencies between frames, making them suitable for videos or image sequences.
LSTM-based models, a type of RNN, capture long-range dependencies in sequential data. \cite{zhang2014LSTM} proposed an LSTM-based method that aligns face landmarks in videos using temporal information.
Cascaded Shape Regression with Temporal Coherence (CSR-TC) \cite{asthana2014incremental} is an extension of cascaded shape regression for video face alignment, CSR-TC improves alignment by incorporating temporal coherence and exploiting shape information from previous frames. The face alignment is perfromed on each frame. 

\textit{Synchronized images from multiple cameras:} In this category, face alignment methods work with images captured simultaneously by multiple cameras. These methods can handle the challenges of pose variations and occlusions more effectively by exploiting the additional spatial information by performing face alignment on each frame.
Multi-view Constrained Local Models (MvCLMs) \cite{zhang2004pose} extend single-view CLMs for multi-view scenarios, MvCLMs jointly optimize local feature detectors across different views, resulting in better alignment performance.
Multi-view Active Appearance Models (MvAAMs)\cite{tzimiropoulos2013optimisation} based on the AAM framework for multi-view scenarios, MvAAMs learn a joint appearance model across different views, allowing simultaneous face alignment in multiple camera settings.
3D Dense Face Alignment (3DDFA) \cite{zhu20163DDFA} is a 3D face alignment method compatible with both single images and multi-view data. In multi-view scenarios, 3DDFA uses a 3D Morphable Model (3DMM) to estimate 3D face landmarks by fitting the model to images from multiple synchronized cameras.

The choice of input data depends on the specific requirements of the application and the trade-off between performance and complexity. Single image methods are simple to implement and suitable for real-time applications. The sequence of image methods takes advantage of the temporal information to improve performance. Synchronized image methods take advantage of spatial information to improve performance, but they are more complex to implement and may not be suitable for real-time applications. All these methods may not perform as well when dealing with challenging situations like large pose variations or occlusions.

\subsection{Based on output representation}
Output representation is an essential aspect of face alignment methods, as it defines the format in which the face information is represented. Different methods may output different representations of the face, such as heatmaps or landmarks/keypoints, and these can be either in 2D or 3D forms. We discuss these in detail in Section \ref{section:face_models}.
\subsection{Based on model formalism}
Model formalism plays a significant role in the design and performance of face alignment methods. Different formalisms can offer various benefits and limitations, depending on the problem at hand. 

\textit{Artificial Neural Networks (ANNs)} are general function approximators consisting of a weighted sum of input data fed through a non-linear activation function. For example, convolutional neural networks (CNNs), recurrent neural networks (RNNs), and generative adversary networks (GANs) are examples of face alignment. Deep Alignment Network (DAN) \cite{kowalski2017deep} is a CNN-based method that learns to predict face landmarks from coarse to fine. It uses a stacked architecture of CNNs to gradually refine the landmark predictions.  Hourglass Networks \cite{newell2016stacked} are deep learning architectures that can be adapted for face alignment. They output heatmaps representing the locations of face landmarks. Deep Regression Forests (DRF) \cite{Bulat2017} deep learning-based method that combines the merits of both CNNs and regression forests. It learns to estimate 3D face landmarks and has shown strong performance in face alignment tasks. These methods provide high accuracy and robustness to variations in lighting, pose, and occlusion, but require extensive data and computational resources for training.

\textit{Graphical Models} represent relationships between variables using graphs and can be used for face alignment tasks. Active Shape Models (ASM) \cite{Cootes1995} learn shape variations from a training set and iteratively update shape parameters to minimize the difference between predicted and observed local features. Constrained Local Models (CLMs) \cite{Cristinacce2006CLM} consist of local feature detectors, typically learned using regression techniques. They work by iteratively updating the shape parameters to minimize an objective function that measures the consistency between the predicted and observed local features. Graphical models offer a balance between performance and complexity, but might struggle with large pose variations or occlusions compared to deep learning methods.

\textit{Regression-based techniques:} These methods directly estimate face landmark coordinates using regression techniques. Examples include Explicit Shape Regression (ESR) \cite{Cao2012ESR} learns a cascade of regression models to directly estimate the positions of face landmarks from image features. It has shown excellent performance in face alignment tasks. Supervised Descent Method (SDM) \cite{xiong2013SDM} is a generic optimization framework that employs cascaded regression models to minimize the nonlinear least squares objective function. Regression-based techniques are simple to implement and suitable for real-time applications.

\subsection{Based on parameter estimation/optimisation}
Parameter estimation and optimization methods play a crucial role in face alignment, as they determine how model parameters are adjusted to best fit the input data. In some cases, these methods are closely intertwined with model formalism. 

\textit{Gradient descent} is an optimization algorithm used for minimizing a loss function, which is a measure of how well a model is performing on a given task. Starting with a parametric face model, the algorithm predicts the position of facial landmarks. Then it calculates an objective function that quantifies the difference between these predictions and the actual landmarks. Using the gradient of this function, the algorithm determines how to adjust the model parameters to reduce this discrepancy. By continually updating the parameters in the direction of the negative gradient, the model refines its predictions. This process is repeated until the predicted landmarks closely match the actual ones or until other predefined stopping criteria are met.

In the context of deep learning and neural networks, gradient descent is used to adjust the model's parameters (e.g., weights and biases) to minimize the loss function and improve the model's performance. The choice of parameter estimation/optimisation method depends on the specific requirements of the application and the trade-off between performance and complexity. For example, Stacked Hourglass Networks \cite{newell2016stacked} and RetinaFace \cite{deng2020retinaface} are widely used and effective in many applications but may get stuck in local minima and require careful tuning of the learning rate.

Ultimately, the selection of a parameter estimation and optimization method for face alignment depends on assessment of the specific requirements of the application and a thorough understanding of each method's strengths and weaknesses. By considering factors such as computational requirements, convergence properties, and robustness, one can choose the most appropriate optimization method for the desired application and achieve the best possible performance.

\section{Comparison of face alignment methods based on output face representation} \label{section:face_models}
\begin{figure}[htb]
      \centering
          \includegraphics[width=1\columnwidth]{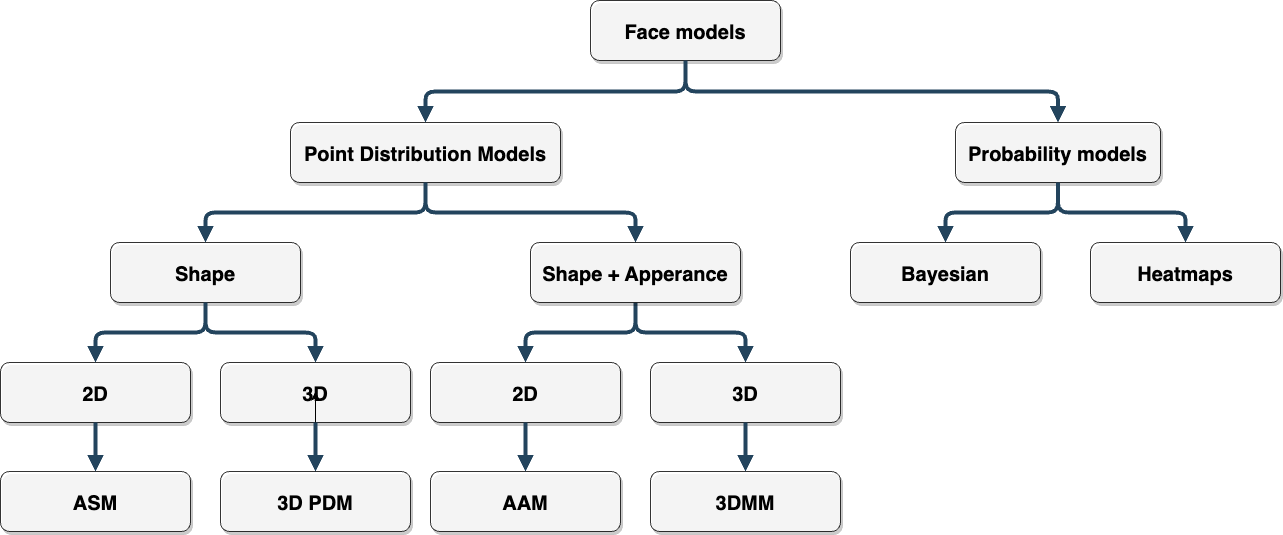}
      \caption{Classification of the face models used in face alignment}
      \label{fig:face_models}
\end{figure}

In this section, we discuss the different face models used in face alignment. Fig. \ref{fig:face_models} presents a categorisation of the different face models based on the way a face model is described mathematically, i.e., as a point landmark model or a probability distribution function. 

\subsection{Active Shape Model (ASM)} \label{subsection:ASM}
\subsubsection{Understanding ASM}
\begin{figure}[htb]
      \centering
          \includegraphics[scale=0.3]{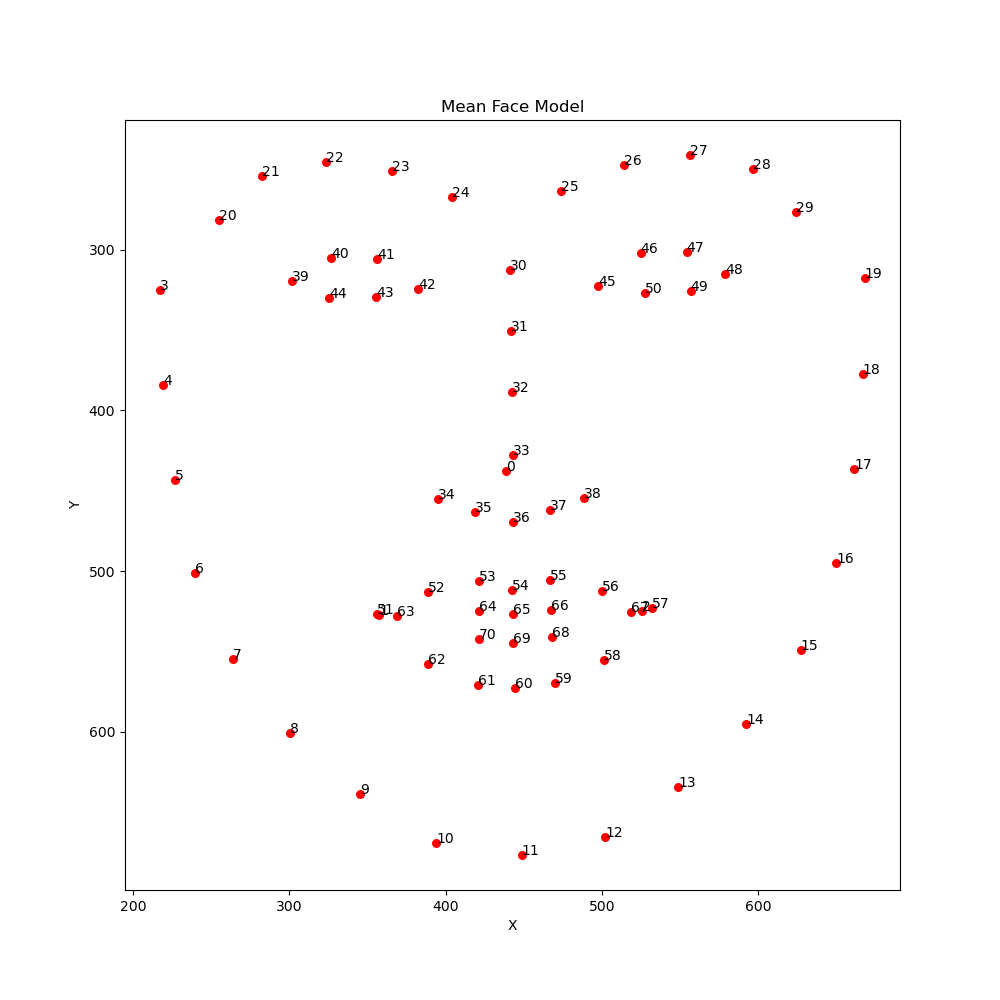}
      \caption{Mean face generated from the training images of the 300W dataset}
      \label{fig:ASM_mean_face}
\end{figure}

Active Shape Models (ASMs) are statistical models that represent the geometric shape of objects using a distribution of landmark points \cite{Cootes1995}. In the context of face alignment, ASMs can model the shape of human faces. The purpose of ASM is to identify and localize landmark points on a face, such as the corners of the eyes, corners of lips or the tip of the nose, etc. 
This is useful in many fields, including computer vision tasks related to face, bio-metrics, medical imaging, etc. 
ASMs capture shape variations and offer robustness to changes in scale, orientation, and position. 
\begin{figure}[htb]
      \centering
          \includegraphics[scale=0.5]{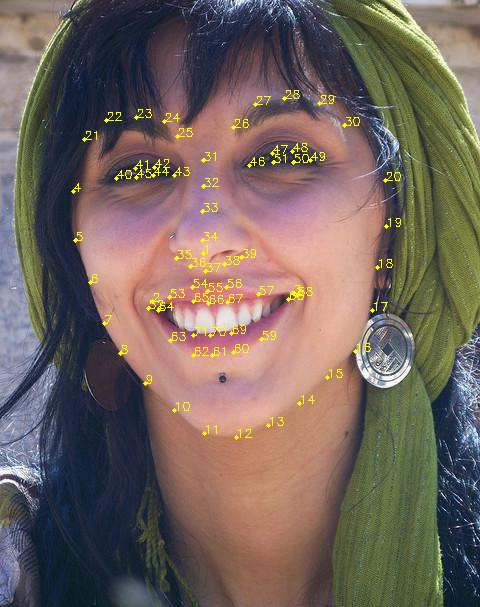}
      \caption{Example of annotated training image from the 300W dataset}
      \label{fig:ASM_training_annotated}
\end{figure}
An ASM model is created by analyzing a training set of images with known landmarks. Fig. \ref{fig:ASM_training_annotated} represents an example of the annotated face in the 300W dataset. 
The landmarks of each training image are aligned and used to create a mean shape as shown in Fig. \ref{fig:ASM_mean_face}. The principal components of the variation from the mean shape are calculated.
One may initiate the process with an assortment of several hundred facial images, which would serve as the training dataset. The training of ASMs requires careful labeling and normalization of the training images. These images could be sourced from various face databases, such as Multi-PIE, LFW, or BioID, HELEN, which provide a wide range of face variations. Labeling is done by marking key landmarks on each face, such as the corners of the eyes, the tip of the nose, and the corners of the mouth, as shown in Fig. \ref{fig:ASM_training_annotated}.
These landmark points are aligned and averaged to produce a mean face shape. Then, the ways that the faces vary from this mean face shape are analyzed, and the principal (most significant) variations are calculated. These principal variations form the basis of the ASM, allowing it to adjust to match different faces.
The basic equation of ASM is the following.
\begin{equation}
\label{Eq:ASM} 
    S = \bar{S} + \sum_{i=1}^{m}b_i P_i
\end{equation}
where $S$ is the shape modeled, $\bar{S}$ is the mean shape calculated from our training set, $P_i$ are the principal components (representing the ways the shape can change), and $b_i$ are the weights or coefficients that control how much of each principal component to include.

\subsubsection{Fitting an ASM model to a new image}
We can use the ASM face model as a template to find the exact locations of the facial features or landmarks in a new image, thus performing face alignment. Assuming that we have the mean face and the most common ways it changes, we can find the best way to align the model with the target face in an image. This involves finding the best values for the coefficients $b_i$ to make the shape of the model $S$ match the landmarks in the new image as closely as possible. A common algorithm used for fitting an ASM to a new image is the Active Shape Model search algorithm, which iteratively adjusts the model to better match the image. It alternates between adjusting the pose (position, scale, and orientation) of the model, adjusting the shape parameters $b_i$, and searching for better matches to the landmarks in the image. The goal of each of these steps is to minimize the difference between the model and the image, thereby achieving an optimal fit. This can also be achieved through methods like a hill-climbing algorithm, which refines the model's fit based on image details. This method requires a well-defined cost function and makes crucial assumptions about the initial shape approximation and the step size, as there is a risk of getting trapped in local optima.
Another method Constrained Local Models (CLMs) determines the accuracy of their fit by combining the prediction errors of the shape model and the local appearance models for each landmark. For effective alignment, it is assumed that the training set can capture the variations in local appearance around each landmark.


\subsubsection{Training an ASM model}
The rest of this section provides a detailed mathematical explanation of the training steps to build the ASM model for a human face. 
To represent the shape of a human face as an object, we need to define a set of landmark points.
Let us represent the shape of a face as a column vector $S_i$, which consists of $x$ and $y$ coordinates of the landmark points, describing the $i$-th shape of the set:
\begin{equation}
\label{eq:2D_shape_coordinates}
    S_i = (x_{i1},y_{i1}, x_{i2}, y_{i2}, \dots, x_{ik}, y_{ik}, \dots, x_{in},y_{in})^T \forall S \in \Re^{2n}
\end{equation}
 where $n$ is the number of landmark points used to describe the face shape, and $(x_{ik},y_{ik})$ are the coordinates of the $k$-th landmark point. Each landmark point represents a particular feature on the face, for example, the corner of the eyes, the tip of the nose, the corner of the lips, etc. Thus, these features need to be consistently labelled from image to image. 
To create a statistical face shape model, we use a set of $N$ training images, each containing a face and its corresponding $n$ landmark points as shown in Fig. \ref{fig:ASM_training_annotated}. However, when dealing with multiple images, where each face appears in a different position, a simple representation becomes inadequate for aligning the faces. 
ASMs employ Procrustes Analysis to learn these variations \cite{gower1975generalized}. Procrustes Analysis is used to align the shapes in the training set by minimizing the dissimilarity between them. This alignment is necessary because it ensures that the statistical shape model is built on consistent and comparable data. The objective of Procrustes Analysis is to find the optimal parameters that minimize the weighted loss function $E_j$, which measures the dissimilarity between shapes. Considering two similar shapes $S_i$ and $S_j$ we account for variations in translation ($t_x, t_y$), rotation ($\theta$), and scaling ($s$), mapping $S_i$ onto $M(s_j,\theta_j)[S_j]-t_j$ so as to minimise the weighted loss function $E_j$. By doing so, we can align the faces in the training images, making it easier to build a reliable shape model.
The weighted loss function $E_j$ is defined as: 
\begin{equation}
    E_j = (S_i - M(s_j,\theta_j)[S_j] - t_j)^T W(S_i M(s_j, \theta_j)[S_j] - t_j)
\end{equation}
where
\begin{equation}
    M(s,\theta) \left[
    \begin{array}{c}
         x_{jk}  \\
         y_{jk}
    \end{array}
    \right]  = \left(
    \begin{array}{c}
        (s \cos \theta) x_{jk} - (s \sin \theta) y_{jk}   \\
        (s \sin \theta) x_{jk} + (s \cos \theta) y_{jk} 
    \end{array} 
    \right)
\end{equation}
operate on each pair $(x_{jk},y_{jk})$ in $S_j$,
\begin{equation}
    t = (t_x, t_y, \dots, t_x, t_y)^T
\end{equation}
contains $n$ copies of $t_x, t_y$ and $W$ is a diagonal matrix of weights of each of the $n$ points. The weights for each point is defined as 
\begin{equation}
    W = \left(\sum_{i=0}^{n} V_{iR_{kl}} \right)^{-1}
\end{equation}
where $R_{kl}$ is the distance between the points $k$ and $l$. Whereas $V_{iR_{kl}}$ is the variance in distance across the entire set of training images.  
When a point moves significantly compared to others, the variance sum is high, and its weight is low. In contrast, stationary points have low variance and high weight and are prioritized to match in different shapes.

ASM is sometimes referred to as a Point Distribution Model (PDM) because one of the fundamental steps in creating an ASM is to model the distribution of landmark points across a set of training images. The goal is to capture the variability and correlations between these points in order to create a statistical shape model. The aim of PDM is to capture variations within dispersed "clouds" of partially correlated landmark points in aligned shapes of the $N$ number of training images. Representing these shapes in a 2n-dimensional space forms a cloud of points within the "Allowable Shape Domain." By assuming the cloud to be approximately ellipsoidal and calculating its center (mean shape) and major axes, new shapes can be generated systematically while navigating the domain. This approach models the cloud's shape in high-dimensional space and captures the relationships between individual landmark positions.
The mean shape $\bar{S}$ for all $N$ training shapes can be calculated as
\begin{equation}
    \bar{S} = \frac{1}{N} \sum_{i=1}^{N} S_i'
\end{equation}
where $S_i'$ represents the aligned shapes normalised based on translation, rotation and scaling. 
We can capture the variability of the shape by computing the covariance matrix of the landmark points. This can be done by subtracting the mean shape from each training shape and computing the covariance. The Covariance Matrix $C$ can be calculated as:
\begin{equation}
    C = \frac{1}{N}  \sum_{i=1}^{N} (S_i'-\bar{S})(S_i'-\bar{S})^T \forall C \in \Re^{2n \times 2n}
\end{equation}
To reduce the dimensionality of the shape variability, we can use Principal Component Analysis (PCA) \cite{wold1987principal}. 
PCA is used to build a statistical model that describes the variations in the training images which are more complex than $s,t$ and $\theta$. 
PCA is a widely used technique in statistics and machine learning for dimensionality reduction, feature extraction, and data compression. In the context of ASM, PCA is used to reduce the dimensionality of shape variability while preserving the most significant modes of variation. This enables the creation of a compact and efficient representation of the shape model, which is crucial when dealing with high-dimensional data, such as images.
The main axes of the face shape are described by the eigenvectors $p_k$ of the covariance matrix (principal components) such that 
\begin{equation}
    C p_k = \lambda_k p_k 
\end{equation}
where $\lambda_k$ is the $k$-th eigenvalue of $C$, and $k \in [1,2n]$.
A shape in the training set can be approximated using a linear combination of the mean shape and weighted sum of the deviations obtained from the first $m$ modes: 
\begin{equation}
    S = \bar{S} + \sum_{i=1}^{m}b_i P_i
\end{equation}
where $\bar{S}$ is the mean shape of the face, $P = (P_1,P_2, \dots, P_m)$ is the matrix of first $m$ eigen vectors, and $b = (b_1, b_2, \dots, b_m)$ is a vector of weights. By varying the parameter $b_k$ within suitable limits, we can make new face shapes similar the training set. Typically, the limit of $b_k$ is
\begin{equation}
    -3\sqrt{\lambda_k} \leq b_k \leq 3\sqrt{\lambda_k},
\end{equation}
since most of the population lies within three standard deviations of the mean. 

\subsubsection{Code implementations}
Table \ref{table:ASM_codes} lists some open-source implementations of ASM that we found. We tried to design our own code for ASM as we wanted to understand the entire process in depth. You can find it in our GitHub repository. We will keep adding additional implementations in it. We also tried Scalismo \cite{sclismo} as they have provided free and paid courses on their website. 
\begin{table}[!ht]
    \centering
    \caption{Code implementations of ASM}
    \label{table:ASM_codes}
    \begin{tabular}{p{7cm}p{1cm}}
    \hline
        Code link & Language \\ \hline
        \url{https://www.mathworks.com/matlabcentral/fileexchange/26706-active-shape-model-asm-and-active-appearance-model-aam} \cite{kroon2023matlab} & MATLAB \\ 
        \url{https://github.com/johnwmillr/ActiveShapeModels} \cite{johnwmiller2017ASM} & MATLAB \\ 
        \url{https://scalismo.org/docs/Tutorials/tutorial13} \cite{sclismo} & Scala \\
        \url{https://github.com/jiapei100/VOSM} \cite{jaipeiVOSM} & C++ \\ 
        \url{https://w3.abdn.ac.uk/clsm/shape/} \cite{shapeASM} & Software \\ 
        \url{http://freesourcecode.net/socialtags/active-shape-model} \cite{freesourcecodeASM}& Varied \\ 
        \url{http://www.milbo.users.sonic.net/stasm/} \cite{Milborrow2014} & C++ \\ 
        \url{https://uomasm.sourceforge.net/} \cite{seshadri2009robust} \cite{milborrow2013multiview} & C++ \\ \hline
    \end{tabular}
\end{table}
\subsubsection{Summary}
A face model represented by an Active Shape Model (ASM) uses a set of landmark points to capture the shape of a face. The model computes the mean shape of the face across training images and captures shape variability using a covariance matrix. Principal Component Analysis (PCA) is applied to reduce dimensionality and represent any face shape as a linear combination of the mean shape and eigenvectors. This makes the ASM model a tool to generate and analyze different face shapes based on the learned shape variations.
While ASMs provide a powerful approach to face alignment, they come with certain assumptions and limitations. For example, the model assumes that shape variations can be adequately captured using a linear combination of a mean shape and a set of principal components. This does not hold for all faces, especially in the presence of rare shapes or extreme facial expressions. The model also assumes that shape variability can be adequately reduced using PCA, which might not be the case with complex or non-linear shape variations. Future research could explore these issues further, developing more robust and versatile models for face alignment.

\subsection{3D Point Distribution Models (3D PDM)}\label{subsection:3D_PDM}
\subsubsection{Understanding 3D PDM}
The 3D point distribution model (PDM) is a statistical model that represents non-rigid shape variations linearly and combines them with a global rigid transformation to place the shape in the 2D image frame \cite{Cootes1995}. \cite{jeni2017dense} has used the PDM to develop a dense face model for real-time face alignment from 2D videos. 
3D PDM is an extension of traditional models to represent three-dimensional shapes and capturing the structural variations of faces. 

By applying Principal Component Analysis (PCA) to a set of aligned 3D shapes, the primary modes of shape variation are identified. This approach forms the basis for the 3D version of Active Shape Models (ASM), allowing the alignment of 3D facial models with both 2D images and 3D scans.
Equation \ref{eq:3DPDM} is a representation of a 3D PDM for projecting the shape onto the image frame. This equation captures the relationship between the 3D shape, its rigid parameters (translation, rotation, and scaling), and the non-rigid shape deformation (using the basis vectors and corresponding weights) to obtain the 3D location of each landmark in the image frame. 

The value of a 3D PDM lies in its ability to create new instances of a facial object by adjusting the parameters that control these transformations. This means that a single model can be adapted to represent a variety of different faces by manipulating the weights applied to the principal components of the shape variation. 
The advantages of using 3D PDM include a more comprehensive representation of faces, increased robustness to varied head poses, realistic facial deformations, and consistency across different camera setups and viewpoints.

\subsubsection{Fitting a 3D PDM to a new image}
For aligning a new face using a 3D PDM, it is necessary to start with a pre-trained model. The task then involves determining the best parameters that minimize the difference between the input face and the model instance. The key challenge in fitting a 3D PDM to a 2D image is handling the added complexity of depth and pose.

Fitting a 3D Point Distribution Model (3D PDM) to a face image involves adjusting the model so that its 2D projection aligns with the facial features in the image. The process begins with an initial guess of the shape and pose of the model. The 3D model is then projected onto the 2D plane and each projected landmark is matched to the closest feature in the image. Based on discrepancies between the landmarks and detected features, the model's parameters are updated. This cycle of projection, matching, and updating iterates until convergence. Throughout the process, regularization ensures the adjusted shape remains plausible, keeping it close to the training set's known shapes. The overarching goal is to capture the face's 3D structure accurately using observed 2D features.

Fitting a 3D Point Distribution Model (3D PDM) to a face image requires specific prerequisites and operates under certain suppositions. A representative set of 3D facial shapes is vital, along with a precise initial placement of the model in the 2D image and local appearance models for guiding alignment. Additionally, regularization techniques are needed to maintain plausible model adjustments. The process presumes that the 3D PDM adequately captures most facial variations, that these variations are linear, and that corresponding landmarks in 3D and 2D denote identical facial features. Furthermore, it typically assumes consistent lighting and texture conditions, the capability to handle both rigid and non-rigid facial deformations, and a specific projection model for the 3D-to-2D transformation. These foundational elements and beliefs guide alignment success and quality.

\subsubsection{Training a 3D PDM}
A PDM is defined as a collection of landmark points in a 3D mesh. Consider a shape model of the face in 3D space, the 3D vertices that make up the mesh can be defined by
\begin{equation}
S = [x_1, y_1, z_1, \dots, x_n, y_n, z_n]^T = [S_1, \dots, S_i, \dots, S_n]^T,
\end{equation}
where $S_i = [x_i; y_i; z_i]$.
The projected shape on the image frame can be defined as
\begin{equation}
\label{eq:3DPDM}
S_i(r, e) = \sum_{i=1}^{n} sR(\bar{S}_i + \Phi_i e) + t
\end{equation}
where $S_i(r, e)$ represents the 3D location of the $i$-th landmark in the image frame after applying the rigid transformation and non-rigid deformation.
$r = [{s, \theta_x, \theta_y, \theta_y, t}]$ are the rigid parameters of the model, $s$ is the global scaling factor, and $R(\theta_x, \theta_y, \theta_z)$ is the rotation matrix obtained by rotating the object around the three axes $(x, y, z)$ by the angles $\theta_x, \theta_y$, and $\theta_z$, respectively.
$t$ is the translation vector applied to the 3D shape.
$\bar{S}_i$ is the mean location of the $i$-th landmark (part of the mean shape $\bar{S}$).
$\Phi = [\Phi_1, \dots, \Phi_n] \in \Re^{3n \times d}$ is the matrix of basis vectors (principal components) obtained from PCA, representing the main modes of shape variation in the 3D PDM.
$e$ is a vector of weights for each principal component, representing the 3D distortion of the face in the linear subspace.
By adjusting the values of $r$ and $e$, one can create new instances of the object with different rigid transformations and nonrigid shape deformations, and project the resulting shape onto the image frame.

\subsubsection{Code implementations}
Table \ref{table:AAM_codes} lists some open-source implementations of AAM that we found. We implemented Mediapipe's framework to perform face alignment and found that it is easy to implement. We tried to find the zface software \cite{jeni2017dense}, however, the link redirects to the Author's personal website and in that we could not find the software.  
\begin{table}[!ht]
    \centering
    \caption{Code implementations of 3D PDM}
    \label{table:3D_PDM_codes}
    \begin{tabular}{p{7cm}p{1cm}}
    \hline
        Code link & Language \\ \hline
        \url{https://developers.google.com/mediapipe} \cite{lugaresi2019mediapipe} & Python\\
        \url{https://github.com/1adrianb/2D-and-3D-face-alignment}\cite{zhu20163DDFA} \cite{Bulat2017} & Python \\
        \url{ http://zface.org} \cite{jeni2017dense} & Software \\
        \url{https://github.com/LeiJiangJNU/DAMDNet} \cite{jiang2019dual} & Python \\
    \hline
    \end{tabular}
\end{table}

\subsubsection{Summary}
The 3D Point Distribution Model (3D PDM) is a statistical tool used in computer vision to represent non-rigid shape variations in a linear fashion. It combines these variations with a global rigid transformation, enabling the placement of the shape into the 2D image frame. The value of a 3D PDM lies in its adaptability, as it can generate a variety of faces by adjusting the parameters governing the transformations.

Fitting a 3D PDM to a new image requires determining the best parameters that minimize the difference between the input face and the model instance. This can be a complex process due to the high dimensionality of the task and is typically performed using optimization algorithms. For example, in their work on real-time face alignment from 2D videos, Jeni et al. \cite{jeni2017dense} employ a 3D PDM as a key component of their method. By adjusting the model's parameters to best match the input face, they are able to accurately align the face in real time, demonstrating the effectiveness of 3D PDMs for face alignment tasks.

Training a 3D PDM involves defining a collection of landmark points in a 3D mesh, and the projected shape onto the image frame is achieved by a combination of scaling, rotation, and translation parameters. Non-rigid shape deformations are controlled by weights applied to the principal components derived from PCA, which represent the main modes of shape variation.

\subsection{Active Appearance Models (AAM)}\label{subsection:AAM}
\subsubsection{Understanding AAM} 
Active Appearance Models (AAMs) are an advanced version of Active Shape Models (ASMs), including both shape and texture for face representation.
In the context of face alignment, AAM combines the shape and texture of a face to create a model that can be used to align and track faces in images and videos \cite{matthews2004activerevisited}\cite{cootes1998active}\cite{cootes2001active}.
Texture is the actual pixel intensity information within the defined shape. For a face, the texture would include the patterns of skin, the colors, any wrinkles, etc. It's the detailed visual information inside the boundary defined by the shape. The combined information of an object's shape and its internal visual details or texture is referred to as the appearance.
An AAM model can effectively represent a wide range of facial appearances by taking advantage of the statistical variation in shape and texture observed in a training set of face images. They are particularly useful in applications that require precise localization of facial features. However, they might struggle with large pose variations or occlusions due to their holistic nature.

\subsubsection{Fitting an AAM model to a new image}
To fit an AAM to a face image, one starts with an initial model position, then adjusts both the shape and texture parameters to closely match the image. The aim is to minimize the difference between the model's appearance and the image. Unlike ASMs, which focus solely on shape by aligning landmarks to image boundaries or features, AAMs provide a more holistic representation by considering the internal visual details (texture) of the face as well. This makes AAMs more comprehensive and potentially more robust to variations like lighting. However, this added complexity means AAMs require more computational effort and can be sensitive to their starting position in the image.

\subsubsection{Training an AAM model}
There are two types of AAMs - independent AAMs that separately model shape and appearance parameters, and combined AAMs that use a single set of linear parameters to characterize both shape and appearance. 

\textit{Independent AAM} models the shape and appearance separately. The shape of the face is represented by a mesh consisting of $n$ vertices, which is denoted by $S$.
The mesh can be defined as:
\begin{equation}
S = (x_1,y_1,x_2,y_2,\ldots ,x_n,y_n)^T
\end{equation}
where $(x_i, y_i)$ represents the coordinates of the $i$-th vertex.
AAMs allow for linear variations in the shape of the face. This means that the shape of the face can be defined as a base shape, denoted by $\bar{S}$, which can be combined linearly with $m$ eigen vectors $P_i$:
\begin{equation}
\label{eq:independent_AAM_shape}
    S=\bar{S} + \sum_{i=1}^{m}b_i P_i.
\end{equation}
where $b_i$ is the shape parameter representing the weights. $\bar{S}$ represents a triangulated base mesh. AAMs are computed using hand-labeled training images and involve applying Principal Component Analysis (PCA) to training meshes. The vectors $\bar{S}$ correspond to the $m$ largest eigenvalues. Before performing PCA, training meshes are usually normalized using Procrustes analysis, which removes variation due to global shape transformations and focuses on local, non-rigid shape deformation. This process, which is similar to ASM, allows the AAM to capture the most significant variations in shape in the training data, which can be used to align and track faces in new images and videos.
For the appearance of an independent AAM, we consider an image $A(q)$ where $q=(x,y)^T$ is a set of pixels on the base shape $\bar{S}$. Thus, appearance can be defined as the mean appearance $\bar{A}(q)$ combined with the $p$ appearance images $A_i(q)$,
\begin{equation}
\label{eq:independent_AAM_appearance}
    A(q) =\bar{A}(q) + \sum_{i=1}^{p} a_i A_i (q) \forall q \in \bar{S}
\end{equation}
where the coefficient $a_i$ is the appearance weight parameter. Table \ref{table:face_models_comparision} represents an example of the appearance of an independent AAM. In the same way as for the shape component, the base appearance $\bar{A}$ and appearance images $A_i$ are calculated by applying PCA to a set of shape-normalized training images. These images are shape-normalized by transforming the hand-labeled training mesh into the base mesh $S_0$, usually employing a piece-wise affine warp and triangulated meshes. The base appearance is represented by the mean image, while the images $A_i$ correspond to the $p$ eigen images associated with the $p$ highest eigenvalues. Performing shape normalization before applying PCA leads to a more condensed appearance eigenspace.

The equations \ref{eq:independent_AAM_shape} and \ref{eq:independent_AAM_appearance} represent the appearance and shape of the AAM, respectively. However, they do not provide an instance of a complete face model, as they are independent of each other, and we need to combine them by warping. Warping in the context of AAMs is the process of transforming the appearance to fit onto a particular shape. In simple terms, think of warping as "stretching" or "molding" the appearance (such as facial features) so that they align correctly with a given shape or mesh, much like draping a printed cloth over a mold. Given the shape parameters $b = (b_1,b_2, \ldots ,b_m )^T$ and the appearance parameters $a = (a_1,a_2, \ldots ,a_p )^T$, we can warp the AAM appearance $A(q)$ to the base shape mesh $s_0$. After affine warping, the 2D image $M$ containing the model instance along with the right shape and size can be defined as 
\begin{equation}
    \label{eq:AAM_warping}
    M(W(q,p))=A(q)
\end{equation}
This signifies that after warping, the destination of a pixel $q$ in $\bar{S}$ is $W(q:p)$. The AAM model $M$ at the pixel $W(q:p)$ in shape $S$ is warped with the appearance $A(q)$. In essence, warping ensures that the appearance details align perfectly with the intended shape, creating a cohesive face model.

\textit{Combined AAMs} on the other hand, link shape and appearance components into a unified model, offering both pros and cons. They provide a more general representation with fewer parameters but lose orthogonality assumptions and limit fitting algorithm choices. Orthogonality ensured that variations in shape were independent of variations in appearance, which simplified the model and fitting process. Without this assumption, combined AAMs often require more sophisticated and computationally intensive fitting algorithms.
The combined AAMs use only one weight parameter $c=(c_1,c_2, \ldots,c_l)^T$ to define both shape and appearance together.
Furthermore, the intertwining can limit the choice of fitting algorithms. In independent AAMs, since shape and appearance were separate, different algorithms optimized for each component could be applied. But in combined AAMs, the algorithm must account for the inter dependencies, which can constrain the choice to those that handle the combined nature effectively.
The shape $S$ is defined as:
\begin{equation}
\label{eq:combined_AAM_shape}
    S=\bar{S} + \sum_{i=1}^{m} c_i P_i
\end{equation}
The appearance $A$ is defined as:

\begin{equation}
\label{eq:combined_AAM_appearance}
    A(q) = \bar{A}(q) + \sum_{i=1}^{p} c_i A_i (q) \forall q\in \bar{S}
\end{equation}

In practice, to balance representation and computational efficiency, the number of parameters $l$ is often chosen to be less than the sum of shape and appearance parameters ($l \leq m + p$). This approach can lead to better fitting by focusing on the most significant modes of variation while discarding the minor ones.
In summary, while combined AAMs offer a more concise representation, they introduce complexities due to the loss of orthogonality between shape and appearance, necessitating careful selection and potentially more sophisticated fitting algorithms.

\subsubsection{Code implementations}
Table \ref{table:AAM_codes} lists some open-source implementations of AAM that we found. We tried to implement the AAM code in MATLAB \cite{kroon2023matlab}. It was easy to understand and implement. We also tried Scalismo \cite{sclismo} as they have provided free and paid courses on their website. 
\begin{table}[!ht]
    \centering
    \caption{Code implementations of AAM}
    \label{table:AAM_codes}
    \begin{tabular}{p{7cm}p{1cm}}
    \hline
        Code link & Language \\ \hline
        \url{https://www.mathworks.com/matlabcentral/fileexchange/26706-active-shape-model-asm-and-active-appearance-model-aam} \cite{kroon2023matlab} & MATLAB \\ 
        \url{https://code.google.com/archive/p/aam-opencv/downloads} \cite{aamopencv}& ~ \\ 
        \url{https://ibug.doc.ic.ac.uk/media/uploads/notebook.html} \cite{ibugAAM} & Python \\ 
        \url{https://uomapm.sourceforge.io/} \cite{seshadri2009robust} \cite{milborrow2013multiview} & C++ \\ 
        \url{http://cvsp.cs.ntua.gr/software/AAMtools/} \cite{papandreou2008adaptive} & C++ \\ 
        \url{https://scalismo.org/docs/Tutorials/tutorial13} \cite{sclismo} & Scala \\ 
        \url{https://github.com/marcovzla/pyaam} \cite{pyaam} & Python \\ 
        \url{https://github.com/phoenix367/AAMToolbox} \cite{aamtoolbox} & C++ \\ 
        \url{https://github.com/htkseason/AAM-Fitting} \cite{htkseasonaam} & Java \\ \hline
    \end{tabular}
\end{table}

\subsubsection{Summary}
Active Appearance Models (AAMs) are statistical tools that are used to represent both the shape and the appearance of an object, such as a human face. The application of AAMs is notably seen in face alignment tasks, where accurate localization of facial features is required.

AAMs can be categorized into two types, Independent AAMs and Combined AAMs. Independent AAMs separately model the shape and appearance of an object. The shape representation in an independent AAM is derived from a base shape that is linearly combined with several eigen vectors, representing the most significant shape variations observed in the training dataset. Similarly, the appearance component of an independent AAM is obtained from a base appearance image, and augmented with significant appearance variations. To generate a complete face model, these shape and appearance components are then warped onto a base shape mesh.

Combined AAMs, on the other hand, link shape and appearance together using a single set of parameters, offering a potential representation with fewer parameters. However, the trade-off lies in the loss of orthogonality assumptions, which can limit the choice of fitting algorithms.

The utility of AAMs is quite broad, offering the ability to represent a wide range of facial appearances, making them useful for tasks requiring precise localization of facial features and face reconstruction. However, their holistic nature may present challenges when dealing with large pose variations or occlusions.
\subsection{3D Morphable Models (3DMM)}\label{subsection:3DMM}
\subsubsection{Understanding 3DMM}
A 3D Morphable Model (3DMM) is a representation of faces in a 3D space that can capture variations in face shape and texture. 
Unlike 2D face models, 3DMMs can handle occlusions and extreme head poses, making them valuable in real-world applications \cite{blanz19993DMM}. 
2D face models assume that the 2D landmarks hidden due to extreme pose or occlusion exist but the exact location is unknown, which leads to poor face alignment. 
In case of a 2D image with extreme head pose or occlusion, the 3D face model assumes that the landmarks exist but all the necessary landmarks are not visible in the image. 
3D Morphable Models have been popular among the computer vision community for 3D face alignment. \cite{egger20203DMMreview} provides a comprehensive review on the history of 3DMMs and also lists all the publicly available 3DMM datasets. The link \url{https://github.com/3d-morphable-models/curated-list-of-awesome-3D-Morphable-Model-software-and-data} provides a list of 3DMM software and datasets. 

\subsubsection{Fitting a  3DMM to a new image}
For aligning faces using a 3DMM, it is necessary to start with an already trained model, such as the Basel Face Model (BFM) \cite{paysan20093d}. To align a new face image, one needs to find the best shape and texture coefficients, as well as camera parameters (like position, focal length, and rotation), that minimize the difference between the projected 3DMM and the 2D image. This is an optimization problem, often solved using techniques such as gradient descent or Levenberg-Marquardt algorithms.

To fit a 3DMM to a face image, an initial model is set, often refined using detected facial landmarks. This model is then projected onto the 2D plane, and differences in appearance between the model and the actual image are calculated. Optimization techniques adjust the model's shape and texture to minimize these differences, iterating until a satisfactory fit is achieved. Assumptions underpinning this process include the model's ability to represent most face variations, consistent lighting across the face, and the capacity to handle both rigid and flexible face movements. In real-world applications, the fitting is enhanced by detecting landmarks, adding constraints to ensure realistic results, using a multi-resolution approach for efficiency, accounting for different lighting conditions, and post-processing for refinement.

In practice, aligning a 3DMM to a 2D image can be more challenging than aligning a 2D model due to the added depth dimension. However, 3DMMs can handle cases of occlusions and extreme poses that are difficult for 2D models. For example, when a face is partially occluded, the 3DMM can still estimate the shape of the occluded part based on the visible parts and the learned 3D face shape distribution.

\subsubsection{Training a 3DMM}
The shape and texture of a 3DMM are represented by vectors $S'$ and $T'$, respectively. The standard procedure to form a 3DMM is to scan real faces and apply PCA to obtain an instance of the 3D face model. 
For example, The Basel Face Model \cite{paysan20093d} was developed using specialized 3D scanning equipment, not just traditional cameras. This scanner employed the structured light technique, projecting specific light patterns onto the face and capturing the deformations of these patterns to derive the face's 3D shape. Additionally, calibrated cameras recorded the face's texture under consistent lighting. Multiple views of the face were captured, especially from the front and sides, to ensure comprehensive coverage and high detail. This approach ensured accurate and detailed 3D representations of facial structures. The BFM is based on 3D scans of 200 individuals, comprising both males and females. This dataset was used to capture a broad range of facial variations, which were essential for creating a comprehensive and versatile 3D Morphable Model.

Consider a 3D face made up of multiple landmark points in the 3D space. The geometry of the 3D face with $n$ vertices can be represented by a vector $S'=(x_1,y_1,z_1,x_2,y_2,z_2,\ldots,x_n,y_n,z_n)^T$ where $S'\in \Re^{3n}$. It contains the $x$, $y$ and $z$ coordinates of the vertices. Assuming that the number of textures is equal to the number of vertices, the vector representing the texture is  $T' =(R_1,G_1,B_1,R_2,G_2,B_2,\ldots,R_n,G_n,B_n)^T$ where $T'\in \Re^{3n}$. It contains the red, blue, and green color values of the $n$ corresponding vertices. Considering the training data containing $N$ number of exemplar faces, the shape of the model can be defined as:
\begin{equation}
\label{eq:3DMM_shape}
    S' = \bar{S} + \sum_{i=1}^{N}\alpha_i S_i
\end{equation}
and the texture of the model can be defined as:
\begin{equation}
\label{eq:3DMM_texture}
    T' = \bar{T} + \sum_{i=1}^{N}\beta_i T_i
\end{equation}
$\Vec{\alpha},\Vec{\beta}\in \Re^{N}$
where $\Vec{\alpha}=(\alpha_1,\alpha_2,\ldots,\alpha_N)^T$ and $\Vec{\beta}=(\beta_1,\beta_2,\ldots,\beta_N)^T$ are the shape and texture coefficients respectively.  
To create a useful face synthesis system, it is crucial to quantify the plausibility of generated faces. This involves estimating the probability distribution for coefficients $\alpha_i$ and $\beta_i$ from a set of example faces, which helps to control the likelihood of the appearance of synthesized faces. Principal Component Analysis (PCA) is used for data compression, which performs a basis transformation to an orthogonal coordinate system formed by the eigenvectors $S_i$ and $T_i$ of the covariance matrices, in decreasing order of their eigenvalues. 
The covariance matrix for shape $C_s$ can be given as:
\begin{equation}
    C_s = \frac{1}{N}  \sum_{i=1}^{N} (S_i'-\bar{S})(S_i'-\bar{S})^T \forall C_s \in \Re^{2N \times 2N} \\
\end{equation}
and for texture $C_t$ can be given as:
\begin{equation}
    C_t = \frac{1}{N}  \sum_{i=1}^{N} (T_i'-\bar{T})(T_i'-\bar{T})^T \forall C_t \in \Re^{2N \times 2N} \\
\end{equation}
where $N$ is the total number of total number of faces in the training data.

The model's ability to represent different faces comes from the fact that $S$ and $T$ are not fixed, but rather can be adjusted using shape and texture coefficients $\alpha_i$ and $\beta_i$, respectively. These coefficients allow the model to generate a variety of faces by linearly combining base shapes and textures learned from a set of example faces. 

\subsubsection{Pre-trained 3DMMs}
\begin{table*}[ht]
\centering
\caption{Comparison of Different 3D Morphable Models}
\begin{tabular}{p{4cm}p{1cm}p{1.5cm}p{1.5cm}p{2cm}p{2cm}p{3cm}}
\hline
\textbf{Name} & \textbf{Year} & \textbf{Training Subjects} & \textbf{Total Images} & \textbf{Vertices per face} & \textbf{Method} & \textbf{Used in Face Alignment} \\
\hline 
FaceScape \cite{yang2020facescape} & 2020 & 938 & 18760 & 2 million (approx.)& 68 view cameras &  \cite{li2021detail}, \cite{wang2022learning}, \cite{wang2021deep}, \cite{guo2023rafare}, \cite{zhuang2022mofanerf}\\ 

Surrey Face Model (SFM) \cite{schoenborn2017surrey} & 2017 & 169 & & 29587 & 3D Scanning & \cite{feng2018evaluation} \\

Large Scale Face Model (LSFM) \cite{booth20163d} & 2016 & 9663 & 10000 & 60000 & 3D Scanning & \cite{gecer2019ganfit} \\

FaceWarehouse \cite{cao2014facewarehouse} & 2014 & 150 & 7050 & 11000 & RGB-D Camera & \cite{jourabloo2016large}, \cite{jourabloo2017pose}  \\

Basel Face Model (BFM) \cite{paysan20093d} & 2009 & 200 & & 53490 & 3D Scanning & \cite{zhu20163DDFA}, \cite{gecer2019ganfit}, \cite{jourabloo2016large}, \cite{jourabloo2017pose}\\

BU-3DFE Database \cite{yin20063d} & 2006 & 100 & 2500 & 10000 & 3D Scanning & \cite{wang2022learning}, \cite{zhang2020unified}, \cite{jan2018accurate}, \cite{feng20183d} \\

FRGC 3D Face Database \cite{phillips2005overview} & 2005 & 688 & 16783 &  & 3D Scanning & \cite{wang2022learning}, \cite{russ20063d},  \\
\hline
\end{tabular}
\label{table:3DMM_comparison}
\end{table*}

There are multiple pre-trained 3DMMs, such as the Basel Face Model (BFM) \cite{paysan20093d}, the Large Scale Face Model (LSFM) \cite{booth20163d}, the Surrey Face Model (SFM) \cite{schoenborn2017surrey}, and the FaceWarehouse model \cite{cao2014facewarehouse}. These models have been widely used for 3D face alignment in various face alignment models. Table \ref{table:3DMM_comparison} presents a list of pre-trained 3D face models that can be used for face alignment. 
For instance, the BFM, developed by the University of Basel, is used in the 3D Dense Face Alignment (3DDFA) method \cite{zhu20163DDFA}. 3DDFA employs a convolutional neural network to regress 3DMM parameters directly from 2D images, allowing 3D face alignment even in cases of extreme poses.

The LSFM is used in the GANFIT method \cite{gecer2019ganfit}, which is a Generative Adversarial Network-based method for 3D face alignment. GANFIT employs a generator network to produce 3DMM coefficients that can reconstruct the input 2D face image, while a discriminator network ensures the generated face shape and texture are realistic.

The Surrey Face Model (SFM), on the other hand, is employed in the 3D Morphable Model Fitting (3DMMF) method \cite{schoenborn2017surrey}. This method uses an iterative fitting strategy to estimate 3DMM parameters from 2D face images.

While these models have proven to be successful, it is important to note that they require a significant amount of computational resources and can be more difficult to interpret and understand compared to traditional methods. However, their ability to learn complex patterns from large amounts of data often leads to superior performance, especially when large amounts of training data are available.

\subsubsection{Code implementations}
Table \ref{table:3DMM_codes} lists some open-source implementations of 3DMM that we found. We tried to implement all of them. We tried to implement all of the 3D face alignment codes provided in \ref{table:3DMM_codes}. Installing the right libraries and dependencies was a challenge as every method uses a unique set of libraries. 
\begin{table}[!htb]
    \centering
    \caption{Code implementations of 3DMM}
    \label{table:3DMM_codes}
    \begin{tabular}{p{7cm}p{1cm}}
    \hline
        Code link & Language \\ \hline
        \url{https://github.com/cleardusk/3DDFA\_V2}\cite{guo20203DDFAV2} & Python \\
        \url{https://github.com/ascust/3DMM-Fitting-Pytorch} \cite{3DMMfitting} & Python \\
        \url{https://github.com/radekd91/emoca}\cite{EMOCA_CVPR2021} & Python \\
        \url{https://github.com/choyingw/SynergyNet}\cite{wu2021synergy} & Python \\
        \url{https://github.com/tranluan/Nonlinear\_Face\_3DMM} \cite{tran2019towards} \cite{tran2018on} \cite{tran2018nonlinear} & Python \\
        \url{https://github.com/menpo/itwmm}\cite{alabort2014menpo} & Python \\
        \url{https://github.com/XgTu/2DASL}\cite{tu2019joint} & Python \\
        \url{https://github.com/LizhenWangT/FaceVerse}\cite{wang2022faceverse} & Python \\
        \url{https://github.com/Microsoft/Deep3DFaceReconstruction}\cite{deng2019accurate} & Python \\
        \url{https://github.com/google/tf\_mesh\_renderer}\cite{Genova_2018_CVPR} & Python \\
        \url{https://github.com/yfeng95/DECA}\cite{feng2021learning} & Python \\ \hline
    \end{tabular}
\end{table}

\subsubsection{Summary}
3D Morphable Models (3DMMs) are a valuable tool in the field of computer vision, providing a comprehensive representation of faces in a 3-dimensional space. The main advantage of 3DMMs over 2D face models is their ability to handle occlusions and extreme head poses, which can pose significant challenges in real-world applications.

Fitting a 3DMM to a 2D image involves finding optimal shape, texture, and camera parameters, solved through optimization algorithms. The 3DMM is trained using PCA on face scans, forming a model that can generate a variety of faces based on shape and texture coefficients. Pretrained 3DMMs like the Basel Face Model and Large Scale Face Model have been employed in 3D face alignment methods, offering superior performance, albeit at the cost of computational resources.

\subsection{Bayesian Mixture Model}\label{subsection:BMM}
\subsubsection{Understanding Bayesian Mixture Model}
The Bayesian Mixture Model is a probabilistic approach to representing faces using a mixture of probability distributions that can align faces from multiple viewpoints \cite{Zhou2005BMM}. 
The goal is to learn a model that can represent different face shapes and the likelihood that certain points are visible or occluded. It uses a two-part model to represent a face:

\textit{Mixture Shape Model:} This model represents the shape of the face using a mixture of probability distributions. It's learned after aligning all facial shapes to a common coordinate frame using a technique called Generalized Procrustes Analysis (GPA). The model consists of multiple clusters, with each cluster having a weight, center, and principal matrix.

\textit{Mixture Visibility Model:} This model represents the visibility of landmark points on the face using a mixture of Bernoulli distributions. The probability of a point being visible in a cluster is computed based on the training data.

This face model addresses the issues of multi-modality and varying feature points by incorporating them into a unified Bayesian framework. It employs a mixture model to represent shape distribution and point visibility, subsequently deriving the posterior of the model parameters based on observation with unknown valid feature points. 

\subsubsection{Fitting a Bayesian Mixture Model to a new image}
Users who want to use BMM for face alignment can leverage the model's ability to accommodate multi-modality and varying feature points within a unified Bayesian framework. The model's shape and visibility components can be used to determine the best fit for a given face, even in cases of partial occlusion or varying viewpoints. 

\subsubsection{Training a Bayesian Mixture Model}
Consider a facial shape consisting of $n$ landmark points. We can represent this shape using a 3D vector $(x,v)$, where $(x_{i1},x_{i2})$ are the coordinates of the $i$th landmark point, and $v_i$ is a binary variable that indicates whether the $i$th point is visible (1) or not (0). The goal is to learn a model that can describe different face shapes and the visibility of landmark points.

To achieve this, we first align all the shapes into a common coordinate frame using Generalized Procrustes Analysis (GPA). Then, we create a 2-cluster mixture model to represent the face shapes:
\begin{equation}
\label{eq:bayesian_location}
\begin{split}  
    p(x|b) = \sum_{i=1}^{n} \pi_i f_i (x|b_{i}) = \\
    \sum_{i=1}^{n} \pi_i (2 \pi \sigma_i^2)^{-n} \exp \{ -\frac{1}{2\sigma_i^2} || x-\mu_{i} -P_{i}b_{i}||^2 \}
\end{split}
\end{equation}
where $m$ represents the cluster number and $\pi_i$ is the cluster weight. $f(.|b_{i})$ is the normal density function where $b$ is the shape parameter and each $b_{i}$ follows the Gaussian distribution $p(b_{i})= n(0,C)_{i}$ in which $C_{i}$ is the covariance matrix. $\mu_{i}$ is the centre of the $i$th cluster and $P_{i}$ is the principal matrix whose columns are the eigenvectors.
To model the visibility of landmark points, we use a mixture Bernoulli model.
\begin{equation}
\label{eq:Bayesian_visibility}
    p(v) = \sum_{i=1}^{m} \pi_i \prod_{k=1}^{N} \left[ (q_k^{(i)})^{v_k} (1- q_k^{(i)})^{1- v_k} \right]
\end{equation}
where $\pi_i$ is similar to the one in equation \ref{eq:bayesian_location} and $q_k^{(i)}$ represents the probability of the $k$th point to be visible for shapes in the $i$th cluster.

\subsubsection{Code implementations}
We could not find any code implementations for the Bayesian Mixture Model. 

\subsubsection{Summary}
The Bayesian Mixture Model (BMM) is a probabilistic method for representing and aligning faces. It consists of two parts: a Mixture Shape Model, which uses a combination of probability distributions to represent face shapes, and a Mixture Visibility Model, which uses Bernoulli distributions to depict the visibility of facial landmarks. By incorporating multi-modality and varying feature points within a unified Bayesian framework, BMM handles varying viewpoints and partial occlusions effectively. Training a BMM involves using a 2-cluster mixture model to represent face shapes and visibility of landmark points, making it versatile for various face recognition tasks.

\subsection{Heatmaps}\label{subsection:Heatmaps}
\subsubsection{Understanding Heatmaps}
Heatmaps are a continuous, two-dimensional representation of face landmark locations in an image, where the intensity at each pixel location $(x, y)$ corresponds to the likelihood or confidence of a face landmark being present at that location. 
In the context of facial alignment, a face model is a representation that captures the spatial structure of facial landmarks. Heatmaps can be considered a face model used for face alignment because they capture this structure by representing the location of each landmark as a 2D Gaussian distribution on a 2D grid. Each heatmap corresponds to a specific landmark, and the intensity of each pixel in the heatmap represents the probability of the landmark being at that location.
Heatmaps provide a differentiable and robust representation of face landmarks, making them particularly suitable for deep learning-based face alignment methods.
\cite{bulat2016pose} was the first work that presented the usage of heatmaps for human head pose estimation. After this, many face alignment papers started to use heat maps too. 
\cite{bulat20163DFAW}, \cite{kowalski2017DAN}, \cite{wang2019Awing}, \cite{lan2021hih} are some examples of deep-learning-based face alignment methods that use heatmaps.

\subsubsection{Fitting process of heatmaps}
Users who want to perform face alignment using heatmaps can use the heatmap values to estimate the likely position of each facial landmark. The highest value in a given heatmap (ideally 1) indicates the most probable location of the associated landmark. Since each heatmap gives a spatial distribution of a particular face landmark, they can be used to estimate the landmarks' locations even in the presence of noise or occlusion.

\subsubsection{Training a deep learning model using a heatmap}
Consider a dataset with $N$ annotated facial images, where each image has the true positions of facial landmarks labeled, such as $S =(x_1, y_1, x_2, y_2, \dots , x_n, y_n)$ for $n$ landmarks.
In the context of face alignment, a heatmap $H(x, y)$ represents the spatial distribution of face landmarks in an image. A common approach to generating heatmaps is to estimate the location of each facial landmark by representing its position as a 2D Gaussian distribution centered around the landmark's true position. 
A heat map $H_k(x, y)$ for the $k$-th landmark can be represented using the 2D Gaussian distribution.
\begin{equation}
\label{eq:heatmaps}
    H_k(x, y) = \exp \left(-\frac{(x - x_k)^2 + (y - y_k)^2}{2 \sigma^2} \right)
\end{equation}
where $H_k(x, y)$ represents the heatmap for the $k$-th landmark at pixel location $(x, y)$, $(x_k, y_k)$ are the coordinates of the $k$-th landmark, and $\sigma$ is a parameter that controls the spatial spread of the heatmap around each landmark. Assuming that the heatmap channels for different landmarks are combined into a single one, $N$ represents the total number of face landmarks. In practice, it is more common to have separate heatmap channels for each face landmark, leading to $N$ individual heatmaps. The range of the heatmap $H(x, y)$ is $[0, 1]$. The value will be the highest ($1$) at the exact location of the face landmark, and it will decrease as the distance from the landmark location increases, approaching $0$ as the distance becomes larger.
When the pixel location $(x, y)$ coincides with the landmark location $(x_i, y_i)$, the exponent term becomes $0$, and the value of $H(x, y)$ is:
$H(x_i, y_i) = \exp(0) = 1$.
As the distance between the pixel location $(x, y)$ and the landmark location $(x_k, y_k)$ increases, the exponent term becomes more negative, causing the value of $H(x, y)$ to decrease towards $0$. Since the exponential function has a lower bound of $0$, the minimum value of $H(x, y)$ is $0$.
There are different types of heatmaps used in face alignment methods, depending on the specific approach and desired properties. For example, in the part-based heatmap, different parts of the face, such as the eyes, nose, and mouth, have their own separate heatmaps. This approach can help provide more focused and accurate landmark localization by modeling the spatial relationships between face parts. \cite{bulat20163DFAW} is an example of a method that uses part-based heatmaps. 

\subsubsection{Code implementations}
Table \ref{table:heatmap_codes} lists some open-source implementations of heatmaps that we found. We implemented all the methods in \ref{table:heatmap_codes}. Installing the right libraries and dependencies was a challenge as every method uses a unique set of libraries. 
\begin{table}[!htb]
    \centering
    \caption{Code implementations of Heatmaps for face alignment}
    \label{table:heatmap_codes}
    \begin{tabular}{p{7cm}p{1cm}}
    \hline
        Code link & Language \\ \hline
        \url{https://github.com/tyshiwo/FHR_alignment} \cite{tai2019towards} & Lua \\
        \url{https://github.com/starhiking/HeatmapInHeatmap}\cite{lan2021hih} & Python \\ 
        \url{https://github.com/protossw512/AdaptiveWingLoss}\cite{wang2019Awing} & Python \\ 
        \url{https://github.com/snow-rgb/Fast-Face-Alignment} \cite{fastfacealignment} & Python \\ 
        \url{https://github.com/1adrianb/face-alignment}\cite{Bulat2017} & Python \\ 
        \url{https://github.com/wangyuan123ac/3DFA-GCN}\cite{wang2022learning} & Python \\ \hline
    \end{tabular}
\end{table}

\subsubsection{Summary}
Heatmaps are a popular method in deep learning-based face alignment. They are two-dimensional continuous representations where each pixel's intensity corresponds to the probability of a face landmark existing at that location. They offer robustness and differentiability, which makes them suitable for various face alignment methods. Researchers use the peak values in the heatmap to determine the most probable location of each facial landmark.

In training a deep learning model, heatmaps can be created using a 2D Gaussian distribution centered around the true position of each facial landmark. The heatmap value is highest (1) at the exact location of the face landmark, decreasing as the distance from the landmark increases. Some methods use part-based heatmaps, creating separate heatmaps for different face parts such as the eyes, nose, and mouth, enhancing the precision of landmark localization.

\begin{table*}[!htb]
\centering
\caption{A list of all the face models along with the equations representing the face. The visual representation of the face models and how they fit to an input image. The face alignment methods that have used the respective face models.}
\label{table:face_models_comparision}
\begin{tabular}{p{1cm}p{6cm}p{2.5cm}p{2.5cm}p{3cm}} 
\toprule
  Model & Equation & Structure & Sample & Use cases \\ \midrule
  ASM \cite{Cootes1995} & $ S = \bar{S} + \sum_{i=1}^{m}b_i P_i$, $S \in \Re^{2n}$ & 
       \begin{minipage}{.3\textwidth}
       \includegraphics[scale=0.1]{Images/ASM_mean_face.png}
        \end{minipage} & 
        \begin{minipage}{.3\textwidth}
       \includegraphics[scale=0.09]{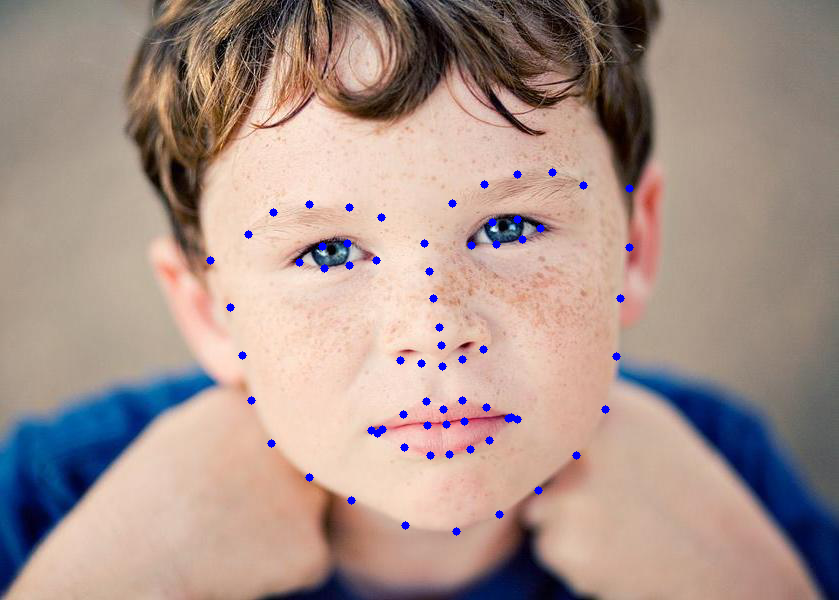}
        \end{minipage} &
        \cite{cootes2000ASM} \\
    3D PDM \cite{Cootes1995} & $S_i(r, e) = \sum_{i=1}^{n} sR(\bar{S}_i + \Phi_i e) + t, \newline S_i \in \Re^{3n}, \Phi_i \in \Re^{3n \times d} $   &
     \begin{minipage}{.3\textwidth}
       \includegraphics[scale=0.25]{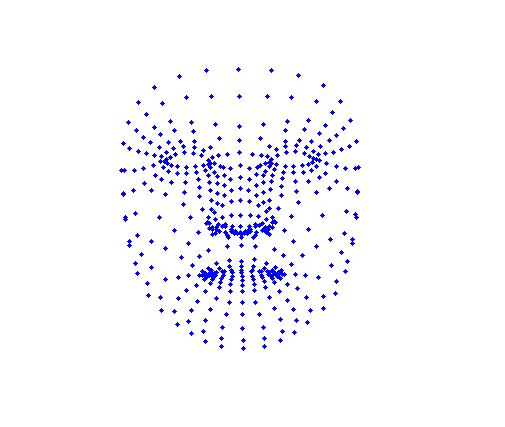}
        \end{minipage} & 
      \begin{minipage}{.3\textwidth}
       \includegraphics[scale=0.10]{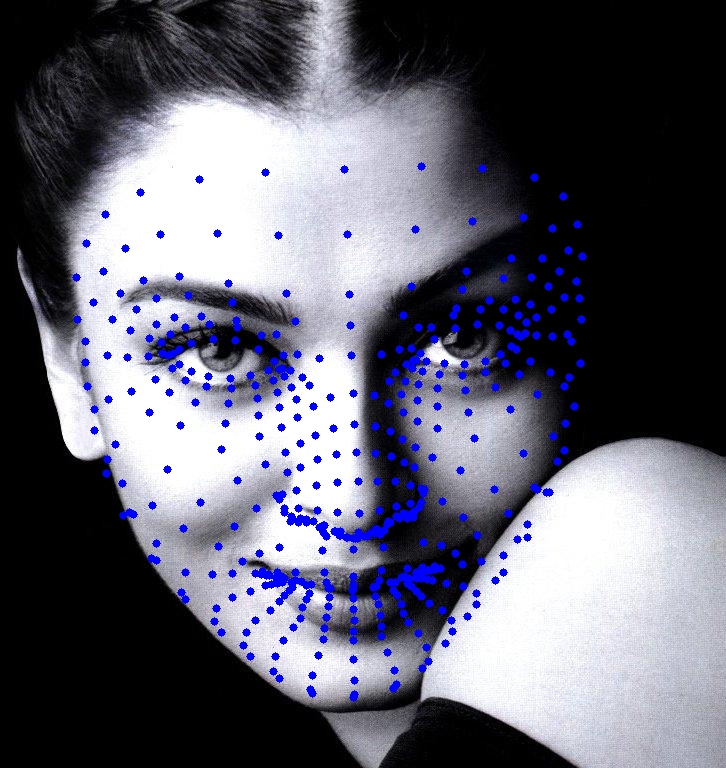}
        \end{minipage} & 
         \cite{jourabloo2015PIFA}, \cite{jeni2017dense}  \\ \\
  AAM \cite{matthews2004activerevisited} &$S=\bar{S} + \sum_{i=1}^{m}b_i P_i$, $S \in \Re^{2n} $ \newline $A(q) =\bar{A}(q) + \sum_{i=1}^{p} a_i A_i (q)$, $A \in \Re^{2n} $ & 
       \begin{minipage}{.3\textwidth}
       \includegraphics[scale=0.13]{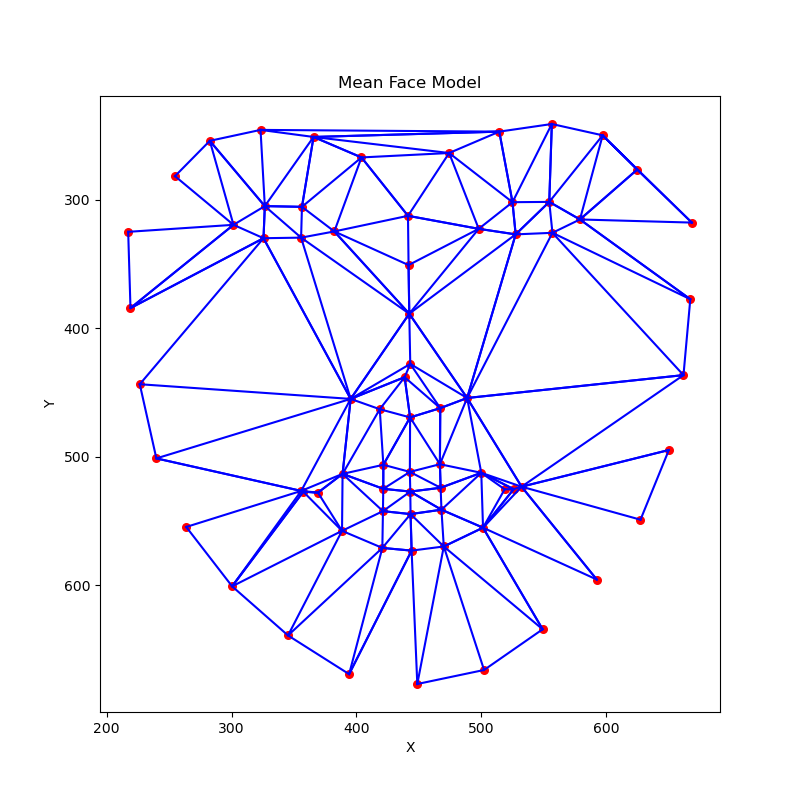}
        \end{minipage} & 
        \begin{minipage}{.3\textwidth}
       \includegraphics[scale=0.12]{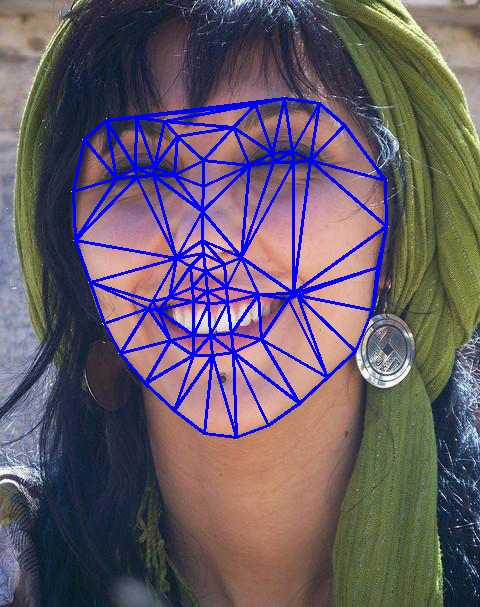}
        \end{minipage} &
        \cite{matthews2004activerevisited} \\

    3DMM \cite{blanz19993DMM} & $ S' = \bar{S} + \sum_{i=1}^{N}\alpha_i S_i$, $S'\in \Re^{3N}$ $\alpha\in \Re^{n}$ \newline  $ T' = \bar{T} + \sum_{i=1}^{N}\beta_i T_i$, $T' \in \Re^{3N}$ $\beta\in \Re^{n}$ &
        \begin{minipage}{.3\textwidth}
       \includegraphics[scale=0.15]{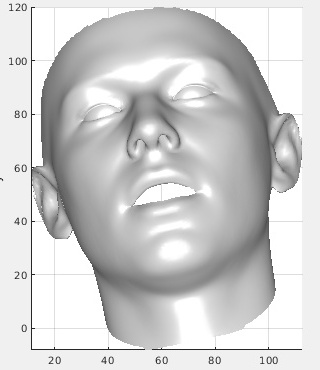}
        \end{minipage} & 
      \begin{minipage}{.3\textwidth}
       \includegraphics[scale=0.09]{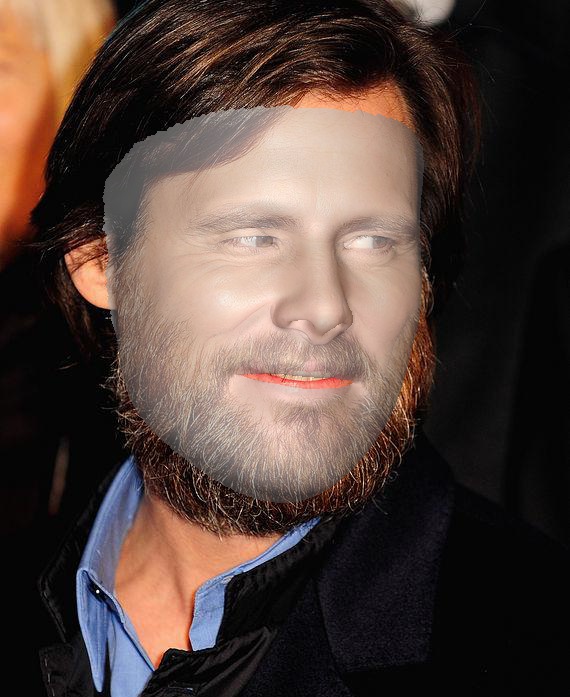}
        \end{minipage} & 
        \cite{zhu20163DDFA}, \cite{feng2018PRNet}, \cite{jourabloo2016PAWF}, \cite{guo20203DDFAV2}, \cite{wu2021synergy}, \cite{jiang2019dual} \\
   
    Bayesian \cite{Zhou2005BMM} & $ p(x|b) =  \sum_{i=1}^{n} \pi_i (2 \pi \sigma_i^2)^{-n} \exp \{ -\frac{1}{2\sigma_i^2} || x-\mu_{i} -P_{i}b_{i}||^2 \}$   &
     \begin{minipage}{.3\textwidth}
       \includegraphics[scale=0.1]{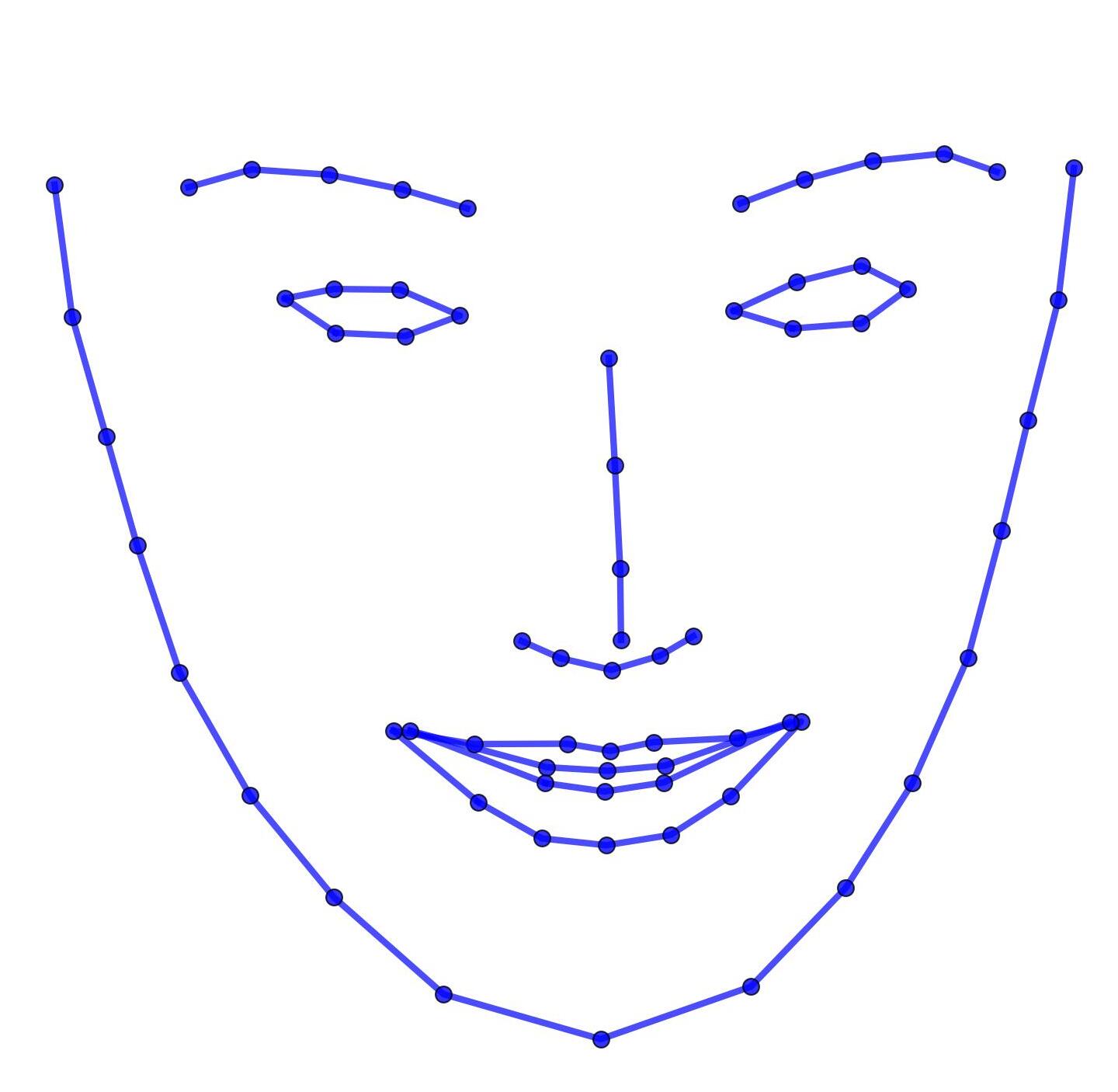}
        \end{minipage} & 
      \begin{minipage}{.3\textwidth}
       \includegraphics[scale=0.06]{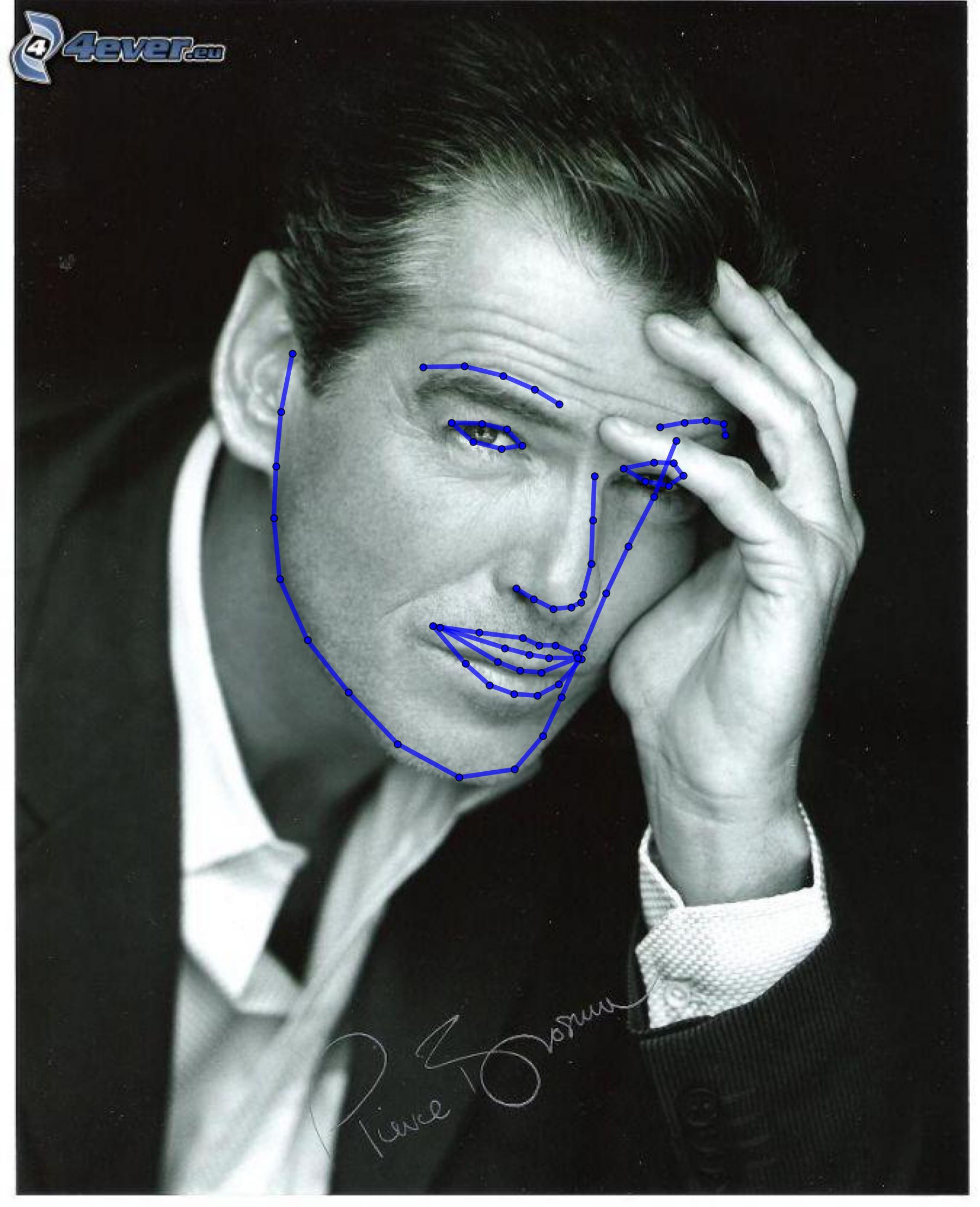}
        \end{minipage} & \cite{Zhou2005BMM}
          \\ \\  
    Heatmaps \cite{bulat2016pose} &  $H_k(x, y) = \exp \left(-\frac{(x - x_k)^2 + (y - y_k)^2}{2 \sigma^2} \right)$ &
     \begin{minipage}{.3\textwidth}
       \includegraphics[scale=0.2]{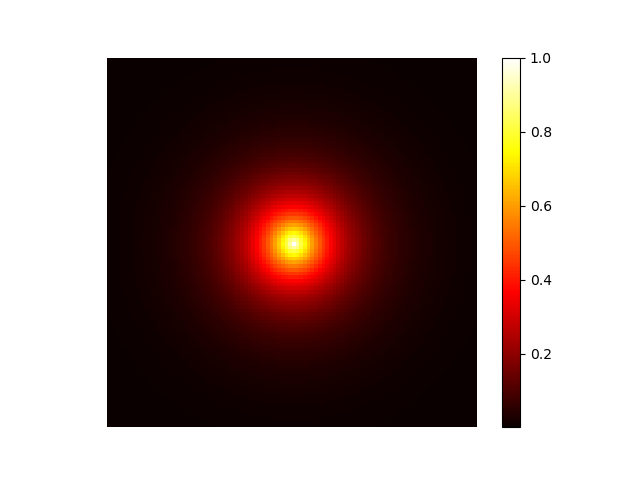}
        \end{minipage} & 
      \begin{minipage}{.3\textwidth}
       \includegraphics[scale=0.2]{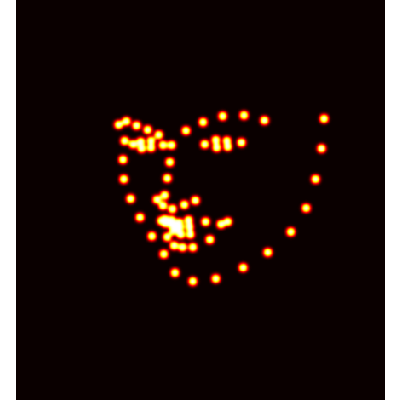}
        \end{minipage} & \cite{kowalski2017DAN}, \cite{wang2019Awing}, \cite{lan2021hih}, \cite{bulat20163DFAW}, \cite{wang2022learning}
          \\ \\
\bottomrule
   
    \end{tabular}
\end{table*}

\section{Discussion} \label{section:Discussion}
In the above section, we broke down each face model and analysed how to train each face model. To gain a perspective on how the face models have evolved over time, the table \ref{table:face_models_comparision} presents a comparison of all the face models. 

The work by \cite{Cootes1995} and \cite{cootes2001active} have set the basis for the formation of the face model. All the Point Distribution face models, that we have discussed in this section, have a similar structure - a mean shape of the face along with some parameters that can be adjusted to fit the model to an input image. The number of parameters vary based on 2D or 3D face model and also varies depending on shape or appearance. Adjusting these parameters to fit the face model to an input image or video is the main goal of face alignment methods. 3DMM, which is commonly used by many researchers for 3D face alignment, has partially solved the problems faced in aligning a face under constrained environmental conditions such as large pose and occlusion. However, there is scope for improving the accuracy for more challenging cases like lower illumination and rapid head movements. Table \ref{table:face_models_comparision} presents all the different face models that have been used in the different face alignment methods. 

Heatmaps, a probabilistic face model, are extensively used in deep learning-based face alignment methods. They represent the spatial structure of facial landmarks as 2D Gaussian distributions on a grid. Each pixel's intensity corresponds to the probability of a landmark being at that location. This probabilistic approach provides a measure of confidence in predicted landmark positions, which can be beneficial for handling uncertainty and improving robustness. However, heatmaps require large amounts of data for training and may struggle with extreme poses or expressions due to the lack of explicit shape constraints.

On the other hand, Active Shape Model (ASM) and Active Appearance Model (AAM) are statistical models that capture the shape and appearance variations in a training set of annotated faces. ASM uses a set of landmark points to represent the shape of a face and models the distribution of these points. AAM extends ASM by also modeling the texture (appearance) inside the delineated shape. These models incorporate explicit shape constraints, which can help handle a wide range of poses and expressions. However, they require careful initialization and may struggle with large variations in lighting, texture, or occlusion.

3D Point Distribution Model (3D PDM) and the 3D Morphable Model (3DMM) extend the concept of ASM and AAM to 3D. They capture the 3D shape and texture variations in a set of annotated 3D face scans. These models can handle pose variations naturally by rotating the 3D model and can provide a 3D face reconstruction from 2D images. However, they require 3D annotated data for training, which is more difficult to obtain, and their computational complexity is higher.

When selecting a face model for face alignment, several factors should be considered:
\begin{enumerate}
    \item Data Availability: If large amounts of annotated 2D images are available, heatmap-based methods can be a good choice due to their strong performance and robustness. If 3D annotated data is available, 3D PDM or 3DMM can be used to leverage the additional 3D information.
    \item Pose Variation: For applications with large pose variations, 3D models (3D PDM or 3DMM) are generally more suitable as they can handle pose variations naturally by rotating the 3D model. Heatmap-based methods can also handle moderate pose variations if trained with sufficient data.
    \item Appearance Variation: If the application involves large variations in lighting, texture, or occlusion, heatmap-based methods may be more robust due to their data-driven nature. However, for applications with specific appearance constraints (e.g., medical imaging), ASM or AAM may be more suitable due to their explicit shape and appearance modeling.
    \item Computational Resources: If computational resources are limited, 2D models (heatmaps, ASM, AAM) are generally more efficient than 3D models (3D PDM, 3DMM).
\end{enumerate}

In summary, the choice of face model for face alignment depends on the specific requirements and constraints of the task. A careful consideration of these factors can help select the most suitable model and achieve the best performance.
\section{Conclusions and Future work} \label{section:Conclusion}
Face alignment methods can be classified in four different ways, namely: input image, output representation, model formalism, and parameter estimation/optimisation. In this review, we focus on an in-depth discussion of the various output face representations used in face alignment methods. These methods can be broadly categorised into statistical face models and probabilistic face models. 
In the second part of this review, we will provide an overview of the face alignment methods based on parameter estimation/optimisation. We will focus more on the recent deep-learning based face alignment methods as they have achieved high accuracy as compared to the previous state-of-the-art methods.
In conclusion, our review highlights the various face models used in face alignment and their strengths and limitations. While these models have proven effective in achieving high accuracy in face alignment, there is still room for improvement. To enhance the accuracy and robustness of face alignment, future research can focus on improving the flexibility and adaptability of face models, particularly for handling extreme pose cases and variations in face expression. By incorporating more diverse and comprehensive data in the training process, face models can become more robust and versatile, leading to further advances in the field of face alignment. 
Next, we list a few thoughts on future directions.

\begin{enumerate}
    \item Incorporation of Temporal Information: Most face alignment methods currently focus on processing static images. However, in many practical applications, such as video processing or real-time face tracking, we have access to a sequence of images over time. This temporal information can provide additional context that can be crucial to accurate face alignment. For example, the movement of facial landmarks from one frame to the next is often smooth and predictable, and this information can be used to improve the accuracy and robustness of facial alignment. Future research could explore the development of face alignment methods that leverage temporal information, for example, by using recurrent neural networks or other types of sequence modeling technique.
    \item Fusion of 2D and 3D Information: While 2D face alignment methods are more computationally efficient, 3D methods can provide more detailed and accurate information about the structure of the face. However, 3D methods often require more complex models and more computational resources. A promising direction for future research is the fusion of 2D and 3D information. This could involve, for example, using a 2D method to provide a rough initial alignment, which is then refined using a 3D model. Alternatively, it could involve developing hybrid models that can process both 2D and 3D information simultaneously. This could potentially lead to face alignment methods that combine the efficiency of 2D methods with the accuracy of 3D methods.
    \item Using segmentation as face alignment: In the context of face alignment, the segmentation approach could potentially provide a more detailed and nuanced representation of facial features, which could be beneficial for handling complex and challenging conditions such as occlusions. However, it is important to note that the segmentation approach may also introduce additional complexities, such as the need for more sophisticated models and algorithms, and the requirement for high-quality, finely annotated training data. Face alignment is affected by facial expressions. There is scope for research into creating morphable expression-aware model. 
\end{enumerate}

Looking ahead, we have identified several promising research directions, including the incorporation of temporal information, the application of explainable AI techniques, and the fusion of 2D and 3D information. We have also highlighted the importance of considering practical applications, challenges, and limitations when developing and selecting face models for face alignment.

As the field of face alignment continues to evolve, we anticipate that new face models and methods will continue to emerge and existing models will be further refined and improved. We hope that this survey will serve as a valuable resource for researchers and practitioners in the field and inspire further exploration and innovation in face alignment.



\appendices
\section{Definitions}
In this section, we provide the definitions of various key terms derived from well-cited articles. 
\subsection{Face alignment}
\cite{jin2017face} - "Fiducial facial points refer to the predefined landmarks on a face graph, which are mainly located around or centered at the facial components such as eyes, mouth, nose and chin. Localizing these facial points, which is also known as face alignment, has recently received significant attention in computer vision, especially during the last decade."

\cite{zhu20163DDFA} - "Face alignment, which fits a face model to an image and extracts the semantic meanings of facial pixels, has been an important topic in CV community.
Traditional face alignment aims to locate face fiducial points like “eye corner”, “nose tip” and “chin center”, based on which the face image can be normalized."

\cite{jourabloo2016large} - "Face alignment is the process of aligning a face image and detecting specific fiducial points, such as eye corners, nose tip, etc.."

\cite{feng2018PRN} - "Face alignment that aims at detecting a special 2D fiducial points is commonly used as a prerequisite for other facial tasks such as face recognition and assists 3D face reconstruction to a
great extent."

\cite{zhu2015CFSS} - "Face alignment aims at locating facial key points automatically. It is essential to many facial analysis tasks,e.g. face verification and recognition, expression recognition, or facial attributes analysis."

\cite{cao2014ESR} - "Face alignment or locating semantic facial landmarks such as eyes, nose, mouth, and chin, is essential for tasks like face recognition, face tracking, face animation, and 3D face modeling."

\subsection{3D Morphable Model}
\cite{egger20203DMMreview} - "A 3D Morphable Face Model is a generative model for face shape and appearance that is based on two key ideas: First, all faces are in dense point-to-point correspondence, which is usually established on a set of example faces in a registration procedure and then maintained throughout any further processing steps. Due to this correspondence, linear combinations of faces may be defined in a meaningful way, producing morphologically realistic faces (morphs).
The second idea is to separate facial shape and color and to disentangle these from external factors, such as illumination and camera parameters. "

\subsection{Search Query}
\begin{enumerate}
    \item Web Of Science - TS=("face alignment" OR "facial landmark detection" OR "facial feature extraction" ) AND TS=("heatmap" OR "Active Shape Model" OR "Active Appearance Model" OR "3D Point Distribution Model" OR "3D Morphable Model" OR "Bayesian mixture model") - 78 hits 
    \item Scopus - TITLE-ABS-KEY ( ( "face alignment" OR "facial landmark detection" OR "facial feature extraction" ) AND ( "heatmap" OR "Active Shape Model" OR "Active Appearance Model" OR "3D Point Distribution Model" OR "3D Morphable Model" OR "Bayesian mixture model" ) ) - 346 hits
    \item "face alignment" OR "facial landmark detection" OR "facial feature extraction"AND " heatmap OR "Active Shape Model" OR "Active Appearance Model" OR "3D Point Distribution Model" OR "3D Morphable Model" OR "Bayesian mixture model" - 47 hits 
\end{enumerate}

\section{Nomenclature}
\begin{table}[!htbp]
\centering
\caption{Nomenclature of mathematical variables used in this survey}
\label{tab:variables_summary}
\begin{tabular}{p{0.8cm}p{1.1cm}p{6cm}}
\hline
Variable & Dimension & Description \\
\hline
$S_i$ & $2n \times 1$ & $i$th shape vector with $n$ landmark coordinates \\

$n$ & scalar & Number of landmark points \\

$N$ & scalar & Number of training images \\

$\bar{S}$ & $2n \times 1$ & Mean shape calculated from the set of training shapes \\

$t_x, t_y$ & scalar & Translation parameters \\

$\theta$ & scalar & Rotation parameter \\

$s$ & scalar & Scaling parameter \\

$E_j$ & scalar & Weighted loss function for Procrustes Analysis \\

$M(s, \theta)$ & scalar & Transformation matrix for scaling and rotation \\

$W$ & $2n \times 2n$ & Diagonal matrix of weights for each landmark point \\

$V_{R_{kl}}$ & scalar & Variance in distance across the entire set of training images \\

$R_{kl}$ & scalar & Distance between points $k$ and $l$ \\

$C$ & $2n \times 2n$ & Covariance matrix of landmark points \\

$p_k$ & $2n \times 1$ & $k$th eigenvector of the covariance matrix \\

$\lambda_k$ & scalar & $k$th eigenvalue of the covariance matrix \\

$P$ & $2n \times m$ & Matrix of first $m$ eigenvectors \\

$b$ & $m \times 1$ & Vector of weights for the first $m$ eigenvectors \\

$b_k$ & scalar & $k$th weight of the weight vector $b$ \\
\hline
$x_i$ & Scalar & $x$-coordinate of the $i$th vertex \\
$y_i$ & Scalar & $y$-coordinate of the $i$th vertex \\
$A(q)$ & Scalar & Appearance of an image at pixel $q$ \\
$q$ & $2 \times 1$ & Set of pixels in the base shape \\
$\bar{A}(q)$ & Scalar & Mean appearance at pixel $q$ \\
$A_i(q)$ & Scalar & Appearance image at pixel $q$ for the $i$th appearance component \\
$a_i$ & Scalar & Appearance weight parameter for the $i$th appearance component \\
$p$ & Scalar & Number of appearance components \\
$M$ & ~ & 2D image containing the model instance \\
$W(q:p)$ & $2 \times 1$ & Warping function \\
$c$ & $l \times 1$ & Combined weight parameter \\
$l$ & Scalar & Total number of shape and appearance components \\
\hline
$S'$ & $3n \times 1$ & 3D shape vector containing $x$, $y$, $z$ coordinates of the $n$ vertices \\
$T'$ & $3n \times 1$ & 3D texture vector contains $R$, $G$, $G$ color values of $n$ corresponding vertices \\
$\bar{S}$ & $3n \times 1$ & Mean shape vector of the 3D PDM \\
$\bar{S}_i$ & $3 \times 1$ & Mean location $(x, y, z)$ of the $i$-th landmark \\
$C_s$ & $2N \times 2N$ & Covariance matrix for shape component \\
$C_t$ & $2N \times 2N$ & Covariance matrix for texture component \\
$R$ & $3 \times 3$ & Rotation matrix \\
$\theta_x, \theta_y, \theta_z$ & Scalar & Angles of rotation around the $x$, $y$, and $z$ axes \\
$s$ & Scalar & Global scaling factor \\
$t$ & $3 \times 1$ & Translation vector \\
$\Phi$ & $3n \times d$ & Matrix of basis vectors (principal components) obtained from PCA \\
$e$ & $d \times 1$ & Vector of weights for each principal component \\
$\alpha$ & $N \times 1$ & Shape coefficients vector for 3DMM \\
$\beta$ & $N \times 1$ & Texture coefficients vector for 3DMM \\
\hline
$x$ & $2N \times 1$ & Coordinates of all landmark points \\

$v$ & $N \times 1$ & Visibility status of all landmark points \\

$\pi_i$ & Scalar & Weight of the $i$-th cluster \\

$\mu{i}$ & $2N \times 1$ & Center of the $i$-th cluster \\

$\sigma_i^2$ & Scalar & Variance for the $i$-th cluster \\

$q_k^{(i)}$ & Scalar & Probability of the $k$-th point being visible in the $i$-th cluster \\
\hline
$\sigma^2$ & scalar & Standard deviation of the Gaussian distribution, controlling the spread of the heatmap \\

$H_i(x, y)$ & scalar & Value at location $(x, y)$ on the heat map for the $i$-th landmark \\
\hline
\end{tabular}

\end{table}

\section*{Acknowledgments}
We thank the Ministry of Science and Technology in Taiwan (107-2634-F-006-009, 108-2218-E-006-046, 108-2634-F-006-009,109-2224-E-006-003, 110-2222-E-006-010) and the National Science and Technology Council in Taiwan (111-2221-E-006-186) for funding this research. As nonnative English speakers, we used ChatGPT 4 by OpenAI to improve our writing and make it more readable. 


\bibliography{bibtex.bib}
%
\bibliographystyle{IEEEtran}


 
\vspace{11pt}

\vspace{11pt}


\vfill

\end{document}